\documentclass[journal]{IEEEtran}

\usepackage{amsmath,amsfonts}
\usepackage{algorithmic}
\usepackage{algorithm}
\usepackage{array}

\usepackage[caption=false,font=normalsize,labelfont=sf,textfont=sf]{}

\usepackage{textcomp}
\usepackage{stfloats}
\usepackage{url}
\usepackage{verbatim}
\usepackage{graphicx}
\usepackage{cite}
\usepackage{soul}
\usepackage{xcolor}
\usepackage{enumitem}
\usepackage{soul}

\usepackage{mathtools}
\newtagform{normalsize}[\normalsize]{\normalsize(}{\normalsize)}

\usepackage{subcaption}
\usepackage{hyperref}

\DeclareMathOperator*{\argmax}{arg\,max}

\begin{document}

\title{Heterogeneous Policy Networks for Composite Robot Team Communication and Coordination}

\author{Esmaeil Seraj$^{1,*}$, ~\IEEEmembership{Student Member,~IEEE}
Rohan Paleja$^{1,*}$, ~\IEEEmembership{Student Member,~IEEE},
Luis Pimentel$^{1}$, 
Kin Man Lee$^{1}$, 
Zheyuan Wang$^{1}$, 
Daniel Martin$^{1}$, 
Matthew Sklar$^{1}$, 
John Zhang$^{2}$, 
Zahi Kakish$^{3}$,
Matthew Gombolay$^{1}$, ~\IEEEmembership{Member,~IEEE}

\thanks{\textcolor{black}{ 
© 2024 IEEE. Personal use of this material is permitted. Permission from IEEE must be obtained for all other uses, in any current or future media, including reprinting/republishing this material for advertising or promotional purposes, creating new collective works, for resale or redistribution to servers or lists, or reuse of any copyrighted component of this work in other works.
The published version of this work appears in \emph{IEEE Transactions on Robotics} and is available at \url{https://doi.org/10.1109/TRO.2024.3431829}.
\\
} \\
$^1$Georgia Institute of Technology, 
$^2$Carnegie Mellon University, 
$^3$Sandia National Laboratories
}
}


\maketitle

\begin{abstract}
High-performing human-human teams learn intelligent and efficient communication and coordination strategies to maximize their joint utility. These teams implicitly understand the different roles of heterogeneous team members and adapt their communication protocols accordingly. Multi-Agent Reinforcement Learning (MARL) has attempted to develop computational methods for synthesizing such joint coordination-communication strategies, but emulating heterogeneous communication patterns across agents with different state, action, and observation spaces has remained a challenge. Without properly modeling agent heterogeneity, as in prior MARL work that leverages homogeneous graph networks, communication becomes less helpful and can even deteriorate the team's performance. In the past, we proposed Heterogeneous Policy Networks (HetNet) to learn efficient and diverse communication models for coordinating cooperative heterogeneous teams. In this extended work, we extend Heterogeneous Policy Networks (HetNet) to support scaling heterogeneous robot teams. Building on heterogeneous graph-attention networks, we show that HetNet not only facilitates learning heterogeneous collaborative policies but also enables end-to-end training for learning highly efficient binarized messaging. Our empirical evaluation shows that HetNet sets a new state of the art in learning coordination and communication strategies for heterogeneous multi-agent teams by achieving an 5.84\% to 707.65\% performance improvement over the next-best baseline across multiple domains while simultaneously achieving a 200$\times$ reduction in the required communication bandwidth.
\end{abstract}

\begin{IEEEkeywords}
Multi-Robot Systems, Heterogeneous Robot Teaming, Neural Graph-Based Communication
\end{IEEEkeywords}

%
\IEEEpeerreviewmaketitle

\section{Introduction}
High-performing human-human teams benefit from communication to build and maintain shared mental models to improve team effectiveness~\cite{macmillan2004communication, mathieu2000influence, seraj2023mixed}. Information sharing is key in building team cognition and enables teammates to cooperate to successfully achieve shared goals~\cite{Salas1992TowardAU, konan2022iterated, mathieu2000influence}. Further, typical communication patterns across human teams widely differ based on the task or role the human assumes~\cite{Taylor2007TheEO}. The field of multi-agent reinforcement learning (MARL) \cite{Busoniu2008ACS} has sought to develop agents that autonomously learn coordination and communication strategies to emulate high-performing human-human teams~\cite{Zhang2013CoordinatingMR, yu2019coordinated, catacora2019cooperative,vinyals2019grandmaster, berner2019dota}. Yet, these approaches have fallen short in properly modeling heterogeneity and communication overhead in teaming~\cite{zhang2021multi, das2018tarmac, du2021learning, mao2017accnet}.

Heterogeneity in robots' design characteristics and their roles are introduced to leverage the relative merits of different \textcolor{black}{robots} and their capabilities~\cite{ravichandar2019strata, seraj2021heterogeneous, seraj2021hierarchical, seraj2019safe, Aragues2021IntermittentCM, seraj2022multi}. We define a heterogeneous robot team as a group of cooperative agents that are capable of performing different tasks and may have access to different sensory information. We categorize \textcolor{black}{robot} agents with similar state, action, and observation spaces in the same \textit{class}. \textcolor{black}{While multi-robot teams can benefit from such heterogeneity through complementarity and combining different capabilities to accomplish complex tasks, communicating is not straightforward as agents do not ``speak'' the same language in such a heterogeneous setting as different classes of agents may need to process received information differently given their context}. We consider scenarios in which agents have different action-spaces and observation inputs from the environment (i.e., due to different sensors) or may not even have access to any observation input (i.e., lack of sensors, broken or low-quality sensors), creating a dependency between \textcolor{black}{sensor-constrained} agents on agents with strong sensing capabilities. As such, in heterogeneous, composite robot teams, communication can be considered a necessity rather than a luxury to improve collaboration performance. 

In our prior work~\cite{Seraj2022LearningED}, we developed an end-to-end heterogeneous graph-attention architecture, Heterogeneous Policy Networks (HetNet), for MARL that facilitates learning efficient, heterogeneous communication protocols among cooperating agents to accomplish a shared task. This framework was able to jointly accomplish modeling heterogeneity within heterogeneous robot teams and learning sender-receiver specific communication patterns. Through this architecture, we achieved a 200$\times$ reduction in the number of communicated bits per round of communication over baselines while also setting a new state-of-the-art (SOTA) in team performance for coordinating composite teams.

In this extension, we are concerned with learning cooperative behaviors for heterogeneous robot teams \textit{at scale}, enabling our approach to learn in large task configurations with sparse reward schemes (i.e., global as opposed to individualized), transfer learned knowledge to novel team compositions and task configurations, and achieve policy robustness to noise-degraded communication channels. 
\textcolor{black}{As we focus on coordinating heterogeneous robot teams across a variety of team compositions, it can become challenging to design complementary and fine-tuned agent-specific reward schemes that lead to high-performing cooperative behavior.} Further, the number of composite team permutations possible is numerous, making \textit{rigid MARL} frameworks that are parametric in the number of agents less useful. For example, if we are able to learn a coordination policy for a heterogeneous team consisting of two autonomous ground vehicles (AGVs) and three unmanned aerial vehicles (UAVs) within a search-and-rescue problem, MARL frameworks that require learning a brand new policy (and communication strategy) for \textit{three} AGVs and \textit{one} UAV can be cost-inefficient. Instead, we require \textit{adaptive} and \textit{scalable} MARL frameworks that can transfer learned knowledge to support heterogeneous robot teams across a variety of configurations. Finally, 
heterogeneity in robot teams naturally leads to differences in communication hardware, such as transmitter or receiver design, which can induce various forms of noise in the communication protocol. This noise can cause erroneous data to be received, ultimately resulting in a degraded policy for decision-making. Scalable communication protocols must be robust to various forms of noise to maximize the utility of each available agent. 

\textcolor{black}{\textit{In this paper, we go beyond our prior work \cite{Seraj2022LearningED} and create a novel MARL framework to support generating coordination policies for heterogeneous robot teams at scale. Beyond this prior work, we first design a Multi-Agent Proximal Policy Optimization variant to improve policy learning feasibility in difficult task configurations, improving both sample efficiency and training speed. Second, we create a scalable architecture that is invariant to the number of \textcolor{black}{robot} agents in each class, the size of the environment, and the size of an agent's observation space, allowing for a MARL framework that can transfer knowledge from smaller task configurations to larger, more difficult configurations. Third, we provide a framework for learning coordination policies under HetNet considering noise-degraded communication channels, addressing an important real-world consideration of multi-agent robot teams.}}
\textcolor{black}{
This manuscript reflects the accepted author version of the paper and accompanies a public release of the codebase, with environment updates incorporated to ensure intended environment behavior and consistency across baseline evaluations.
} 
In this extended version, we present the following contributions:

\color{black}
\begin{enumerate}[leftmargin=*]
     \item We extend HetNet to develop a Multi-Agent Heterogeneous Proximal Policy Optimization (MAH-PPO) framework to learn cooperation policies where agents of a composite team learn inter- and intra-class cooperation policies to efficiently communicate and coordinate their actions with heterogeneous teammates. \textcolor{black}{We find that optimizing with PPO enables us to learn policies in larger-scale domains where learning was inefficient and difficult due to increased heterogeneity within the multi-robot team.} 
     \item We modify our graph-based architecture to include a Fully Convolutional Network (FCN), resulting in invariability to 1) the size of the environment and 2) the number of agents in the team. This facilitates learning a policy that can be \textit{transferred} to large-scale domains, achieving a \textcolor{black}{389.51\%} improvement in sample efficiency and a 80.63\% decrease in run-time required to achieve high-performance. 
    \item We conduct an empirical evaluation across several domains and team compositions. We present evidence that are evolved variant of HetNet sets a new SOTA in learning emergent heterogeneous cooperative policies by achieving at least an 5.84\% to 707.65\% performance improvement over baselines in small domain configurations and a 32.5\% to 1161.35\% performance improvement across larger-scale domain configurations.
    \item We extend our framework to support noisy communication channels and show that HetNet is robust to data perturbations. With a poor signal-to-noise ratio (\textcolor{black}{3.24 db}) under additive white Gaussian noise, HetNet performs just \textcolor{black}{4.57\%} worse than with a perfect communication channel. We introduce two additional formulations for communication loss (range-based and type-based) to allow a more thorough assessment of robustness to noise within a heterogeneous robot team setting.
    \item We demonstrate our algorithm on physical robots on a physical, multi-robot testbed.
\end{enumerate}

In Section \ref{sec:related_work} and \ref{subsec:ProblemFormulation}, we introduce related work and preliminaries, respectively. In Section \ref{sec:method}, we introduce our graph-based architecture that supports modeling heterogeneous teams of various size (contribution 2). In Section \ref{sec:TrainingandExecution}, we describe our training and execution procedure, including our MAH-PPO learning algorithm (contribution 1). In Section \ref{sec:eval}, we describe our evaluation environments, baselines, empirical findings, and ablation study across degraded communication channels (contribution 3 and 5).  Finally, in Section \ref{sec:robo}, we present a real-world robot demonstration (contribution 5) of our algorithm within a heterogeneous robot team. 
\color{black}

\section{Related Work}
\label{sec:related_work}
\noindent\textbf{MARL with Communication} -- Recently, the use of communication in MARL has been shown \textcolor{black}{to enhance} the collective \textcolor{black}{performance} of learning agents in cooperative MARL problems~\cite{singh2018learning, zhang2021multi, das2018tarmac, sukhbaatar2016learning, liu2019multi, Jiang2018LearningAC, Zhang2013CoordinatingMR, chu2019multi, xu2021stigmergic, seraj2023embodied, pesce2020improving, seraj2020coordinated}. DIAL~\cite{foerster2016learning} and CommNet~\cite{sukhbaatar2016learning} displayed the capability to learn discrete and continuous communication vectors, respectively. While DIAL considers the limited-bandwidth problem, neither of these approaches are readily applicable to composite teams or capable of performing attentional communication. TarMAC~\cite{das2018tarmac} achieves targeted communication through an attention mechanism which improves performance compared to prior work. Nevertheless, TarMAC requires high-bandwidth message-passing channels, and its architecture is reported to perform poorly in capturing the topology of interaction~\cite{liu2019multi}. SchedNet~\cite{kim2019learning} explicitly addresses the bandwidth-related concerns. However, in SchedNet, agents learn how to schedule themselves for accessing the communication channel rather than learning the communication protocols from scratch. \textcolor{black}{Furthermore, across several MARL works discussed in this section, including \cite{sukhbaatar2016learning,niu2021multi,das2018tarmac, singh2018learning}, a key limitation is that these frameworks required an individualized reward scheme to scale to larger task configurations. Individualized reward schemes, while assisting with the credit assignment problem in MARL \cite{foerster2018counterfactual}, can learn suboptimal coordination strategies in heterogeneous robot teams. A more effective, albeit more difficult-to-learn strategy is to use a shared team reward scheme, enabling agents to learn cohesive policies that achieve the composite team's goal, rather than learning policies that benefit the skill of a particular class of agent.}
Furthermore, in our MARL approach, we explicitly address the heterogeneous communication problem where agents learn diverse communication protocols and intermediate language representations to use among themselves for cooperation. Our model enables agents to perform attentional communication and send limited-length \textcolor{black}{digitized} messages through class-specific \textcolor{black}{encoder-decoder} channels, addressing the limited-bandwidth issues. 

\textbf{MARL with Graph Neural Networks (GNN)} -- Prior work on MARL has sought to utilize GNNs to model a communication structure among agents \cite{adjodah2019communication}. \textcolor{black}{Deep Graph Network (DGN)~\cite{Jiang2020GraphCR} represents} dynamic multi-agent interaction as a graph convolution to learn cooperative behaviors. This seminal work in MARL demonstrates that a graph-based representation substantially improves performance. In \cite{sheng2020learning}, an effective communication topology is proposed by using hierarchical GNNs to propagate messages among groups and agents. G2ANet~\cite{liu2019multi} proposed a game abstraction method combining a hard and a soft-attention mechanism to dynamically learn interactions between agents. More recently, MAGIC~\cite{niu2021multi} introduced a scalable, attentional communication model for learning a \textcolor{black}{centralized} scheduler to determine when to communicate and how to process messages through graph-attention networks. While \textcolor{black}{these} prior work have successfully modeled multi-agent interactions, they are not designed to address heterogeneous teams directly. HetNet, on the other hand, is designed to capture the heterogeneity among agents and learn an efficient shared language across agents with different action and observation spaces to improve cooperativity.

\textbf{Heterogeneity in Multi-agent Systems} -- In~\cite{multi_agent_communication_hetero}, several types of heterogeneity induced by agents of different capabilities are discussed. \textcolor{black}{As opposed to homogeneous teams, the diversity among agents in heterogeneous teams makes it challenging to hand-design intelligent communication protocols}. In~\cite{xiong}, a control scheme is hand-designed for a heterogeneous multi-agent system by modeling the interaction as a leader-follower system. More recently, HMAGQ-Net~\cite{meneghetti2020towards} utilized GNNs and Deep Deterministic Q-network (DDQN) to facilitate coordination among heterogeneous agents \textcolor{black}{(\textcolor{black}{i.e., those with} different state and action spaces)}. \textcolor{black}{Going beyond this prior work, we build our HetNet model based upon an} actor-critic framework and generalize the problem formulation for \textcolor{black}{state-, action- and observation-space} heterogeneities. Moreover, HetNet facilitates learning efficient binary representations of states as an intermediate language among agents of different types to improve cooperativity.

\textbf{Noisy Communication in Multi-agent Systems} -- \textcolor{black}{Prior work investigating the effects of noise in MARL has primarily focused on noise within the environment~\cite{luo2020fault, kilinc2018multiagent}, where observations may not accurately represent the true state of the environment. While these works demonstrate methods to handle errors from noisy agent observations, they assume a perfect communication channel between the agents. \cite{WU2011487} and \cite{karabag2022planning} propose planning frameworks for multi-agent systems with intermittent or full loss of communication but do not consider noisy data in the communication channel. MARL frameworks such as~\cite{Freed_Sartoretti_Hu_Choset_2020, mostaani2019learningbased, tung2021noisychannel} explicitly model communication noise through a binary symmetric channel or additive Gaussian noise, but none of them address both heterogeneity and bandwidth limitations simultaneously. HetNet, in contrast, can learn \textcolor{black}{near-optimal} strategies under noisy and bandwidth-limited communication for heterogeneous teams, traits that are required for robot teams as they begin to scale.}

\section{Preliminaries}
\label{subsec:ProblemFormulation}

\noindent Founding on a standard Partially Observable MDP (POMDP) \cite{Kaelbling1998PlanningAA}, we formulate a new problem setup termed as Multi-Agent Heterogeneous POMDP (MAH-POMDP), which can be represented by a 9-tuple $\langle \mathcal{C}, \mathcal{N}, \{\mathcal{S}^{(i)}\}_{i\in\mathcal{C}}, \{\mathcal{A}^{(i)}\}_{i\in\mathcal{C}}, \{\Omega^{(i)}\}_{i\in\mathcal{C}}, \{\mathcal{O}^(i)\}_{i\in\mathcal{C}}, r, \mathcal{T}, \gamma \rangle$. $\mathcal{C}$ is a set of all available agent classes in the composite robot team, and the index $i\in\mathcal{C}$ shows the agent class. $\mathcal{N}=\sum_{\langle i\in\mathcal{C}\rangle}N^{(i)}$ is the total number of collaborating agents where $N^{(i)}$ represents the number of agents in class $i$. $\{\mathcal{S}^{(i)}\}_{i\in\mathcal{C}}$ is a discrete joint set of state-spaces. For each class-dependent state-space, $\mathcal{S}^{(i)}$, we have $\mathcal{S}^{(i)}=\left[s_t^{i_1}, s_t^{i_2}, \cdots, s_t^{i_{N^{(i)}}}\right]$, where $s_t^{i_j}$ represents the state-vector of \textcolor{black}{robot} agent $j$ of the $i$-th class, at time $t$. $\{\mathcal{A}^{(i)}\}_{i\in\mathcal{C}}$, is a discrete joint set of action-spaces. For each state-dependent action-space, $\mathcal{A}^{(i)}$, we have $\mathcal{A}^{(i)}=\left[a_t^{i_1}, a_t^{i_2}, \cdots, a_t^{i_{N^{(i)}}}\right]$, forming the joint action-vector of agents of class $i$ at time $t$. $\{\Omega^{(i)}\}_{i\in\mathcal{C}}$ is similarly defined as the joint set of observation-spaces, including class-specific observations. $\gamma\in[0, 1)$ is the temporal discount factor for each unit of time, and $\mathcal{T}$ is the state transition probability density function.

At each timestep, $t$, each agent~\footnote{\textcolor{black}{By ‘agent’ we are mainly referring to a ‘robot agent’.}}, $j$, of the $i$-th class can receive (if the observation input is enabled for class $i$) a partial observation $o_t^{i_j}\in\Omega^{(i)}$ according to some class-specific observation function $\{\mathcal{O}^{(i)}\}_{i\in\mathcal{C}}: o^{i_j}_t\sim\mathcal{O}^{(i)}(\cdot|\bar{s})$. If the environment observation is not available for agents of class $i$, agents in the respective class will not receive any input from the environment (\textcolor{black}{e.g.,} lack of sensory inputs). Regardless of receiving an observation or not, at each time, $t$, each \textcolor{black}{robot} agent, $j$, of class $i$, takes an action, $a_t^{i_j}$, forming a joint action vector $\bar{a}=\left(a_t^{1_1}, a_t^{1_2}, \cdots, a_t^{i_1}, \cdots, a_t^{i_j}\right)$. When agents take the joint action $\bar{a}$, in the joint state $\bar{s}$ and depending on the next joint state, they receive an immediate reward, $r(\bar{s}, \bar{a})\in\mathbb{R}$, shared by all agents and regardless of their classes. \textcolor{black}{Our objective in Equation (\ref{eq:expected_reward}) is to learn \textcolor{black}{near-optimal} policies, per existing agent-class, to solve the MAH-POMDP by maximizing the total expected, discounted reward accumulated by agents over an infinite horizon, i.e., 
\begin{align}
\label{eq:expected_reward}
\pi(\bar{s}) = \argmax_{\pi(\bar{s})\in \Pi}\mathbb{E}_{\pi(\bar{s})}\left[\sum_{k=0}^{\infty}\gamma^kr_{t+k}|\pi(\bar{s})\right] 
\end{align}
}
\vspace{-5mm}
\subsection{\textcolor{black}{Policy Gradient Methods}}
\label{subsec:policy_gradient_methods}
\textcolor{black}{
\noindent Policy Gradient methods are an approach to RL that utilize function approximation, in which each agent $j$ has a policy, $\pi^j_{\theta}(a|s)$, parameterized by $\theta$, that specifies which action, $a$, to take in each state, $s$, to maximize the expected future discounted reward. Policy gradient methods apply gradient ascent to the actor's parameters, $\theta$, based on a gradient estimate of the expected return in Equation \ref{eq:expected_reward}. By the policy gradient theorem~\cite{sutton2018reinforcement}, the expected reward maximization, $J(\theta)$, is maximized via Equation~\ref{eq:policy_grad}, where $a_t^j$ and $o_t^j$ are the action and observation of agent $j$, respectively.} 
\textcolor{black}{
\begin{align}
\label{eq:policy_grad}
    \nabla_\theta J(\theta) = \mathbb{E}_{\pi_\theta^j}\left[\nabla_\theta \log\pi_\theta^j(a_t^j|o_t^j)\hat{A}_t(o_t^j, a_t^j)\right]
\end{align}}
\noindent\textcolor{black}{The \emph{advantage} function, $\hat{A}_t(o_t^j, a_t^j) = Q(o_t^j, a_t^j) - V^{\phi}(o_t^j)$, measures how much better the action is relative to the default action of the current policy \cite{schulman2015high}, where  $Q(o_t^j, a_t^j)$ is the state-action value function approximated by the total discounted rewards, and $ V^{\phi}(o_t^j)$ is the state-value function, approximated according to a critic network parametrized by $\phi$.}

\subsection{\textcolor{black}{Proximal Policy Optimization}}
Proximal Policy Optimization (PPO) was introduced to improve training stability, sample efficiency, and performance in single-agent RL tasks \cite{schulman2015high}. PPO achieves better training stability by constraining the size of gradient updates to the policy, such that the current policy does not change too much from the previous policy. This is achieved by the Clipped Surrogate Objective in Equation (\ref{eq:ppo_clip}), where $r_t(\theta)$ is the probability ratio between the current and old policy before updating (\ref{eq:ppo_ratio}), and $\epsilon$ is a hyper-parameter defining the clip range.

\begin{equation}
    \small
    J^{\text{CLIP}}(\theta) = \mathbb{E}\left[\min \big(r_t(\theta)\hat{A}_t, \text{clip}(r_t(\theta), 1 -\epsilon, 1 + \epsilon)\hat{A}_t\big )\right] \label{eq:ppo_clip}
\end{equation} 
\normalsize
\begin{equation}
r_t(\theta) = \frac{\pi_{\theta}(a_t | s_t)}{\pi_{\theta_{\text{old}}}(a_t|s_t)} \label{eq:ppo_ratio}
\end{equation}
Additionally, an entropy term, $J^{\text{S}}(\theta) = c_1 S(\pi_{\theta}(s_t))$, is added to ensure sufficient exploration, where S is the policy entropy and $c_1$ is the entropy coefficient hyperparameter. PPO also performs multiple steps of mini-batch gradient updates, leading to higher sample efficiency.

\subsection{Graph Neural Networks}
\label{subsec:GNN}
\noindent Graph Neural Networks (GNNs) are a class of deep neural networks that capture the structural dependency among nodes of a graph via message-passing between the nodes, \textcolor{black}{where} each node aggregates feature vectors of its neighbors to compute a new feature vector~\cite{zhou2020graph, Wu2020ACS, Jiang2020GraphCR}. \textcolor{black}{The canonical} feature update procedure via graph convolution operator \textcolor{black}{can be shown as $\bar{h}_j^\prime = \sigma\left(\sum_{k\in N(j)} \frac{1}{c_{jk}}\omega\bar{h}_k\right)$,} where $\bar{h}_j^\prime$ is the updated feature vector for node $j$, $\sigma(.)$ is the activation function and, $\omega$ represents the learnable weights. $k\in N(j)$ includes the immediate neighbors of node $j$ where $k$ is the index of neighbor, and $c_{jk}$ is the normalization term which depends on the graph structure. A common choice of $c_{jk}$ is $\sqrt{|N(j)N(k)|}$. \textcolor{black}{In an $L$-layer aggregation,} a node $j$'s representation captures the structural information within the nodes that are reachable from $j$ in $L$ hops or fewer. However, the fact that $c_{jk}$ is structure-dependent can \textcolor{black}{impair} generalizability of \textcolor{black}{GNNs when scaling the graph's size.} \textcolor{black}{Thus,} a direct improvement is to replace $c_{jk}$ with attention coefficients, $\alpha_{jk}$, computed via \textcolor{black}{Equation}~\ref{eq:attention}. In \textcolor{black}{Equation}~\ref{eq:attention}, $\bar{W}_{att}$ is the learnable weight, $\mathbin\Vert$ represents concatenation, and $\sigma^\prime(.)$ is the LeakyReLU nonlinearity. The Softmax function is used to normalize the coefficients across all neighbors $k$, enabling feature-dependent and structure-free normalization~\cite{velickovic2018graph, wang2020learningRAL}.
\begin{equation}
\label{eq:attention}
\alpha_{jk} = \mathrm{softmax}_k\left(\sigma^\prime\left(\bar{W}_{att}^T\left[\omega\bar{h}_j \mathbin\Vert\omega\bar{h}_k\right]\right)\right)
\end{equation}

\section{Method}
\label{sec:method}
\noindent In this section, we first present an overview of the communication problems and constraints considered in our work. We then describe how to construct a heterogeneous graph given a problem state and present the building block layer of our model, which we refer to as the Heterogeneous Graph-Attention (HetGAT) layer. Next, we develop a novel preprocessing module that enables our architecture to be invariant to the number of \textcolor{black}{robot} agents in each class, the size of the environment, and the size of an agent's observation space. Afterward, we discuss our \textcolor{black}{binarized} encoder-decoder communication channel that accounts for the heterogeneity of messages passed among agents while digitizing messages for effective utilization of bandwidth. Finally, we cover the logistics of utilizing HetGAT layers to assemble our heterogeneous policy network, HetNet, of arbitrary depth.

\subsection{Communication Problem Overview}
\label{sec:CommunicationProblemandOverview}
In this work, we are concerned with the problem of coordinating a robot team via fostering direct communication among interacting agents. We consider MARL problems wherein multiple agents interact in a single environment to accomplish a task that is of a cooperative nature. We are particularly interested in scenarios in which the agents are heterogeneous in their capabilities, meaning agents can have different state, action, and observation spaces in forming a composite team. To collaborate effectively, agents must share messages that express their observations under a Centralized Training and Distributed Execution (CTDE) paradigm~\cite{foerster2018counterfactual, kim2019learning}.

In learning an end-to-end communication model, we take a series of problems and constraints into consideration: (1) heterogeneous messages, where agents of different classes have different action and observation spaces, resulting in different interpretations of sent/received messages; (2) Attentional and scalable communication protocols, such that agents incorporate attention coefficients depending on the agent/class they are communicating with for coordinating with teammates in any arbitrary team sizes; \textcolor{black}{(3) \textcolor{black}{Learning communication models for Low-Size, -Weight, and -Power (Low-SWAP) systems, where due to limited communication bandwidth, \textcolor{black}{robot} agents must learn to communicate in a highly efficient} shared intermediate ``language'' (\textcolor{black}{e.g.,} limited-length binarized messages); (4) Limited-range communications, where agents can only exchange messages when they are within close proximity.}

\begin{figure*}[t!]
	\centering
	\includegraphics[height=8.75cm, width=2\columnwidth]{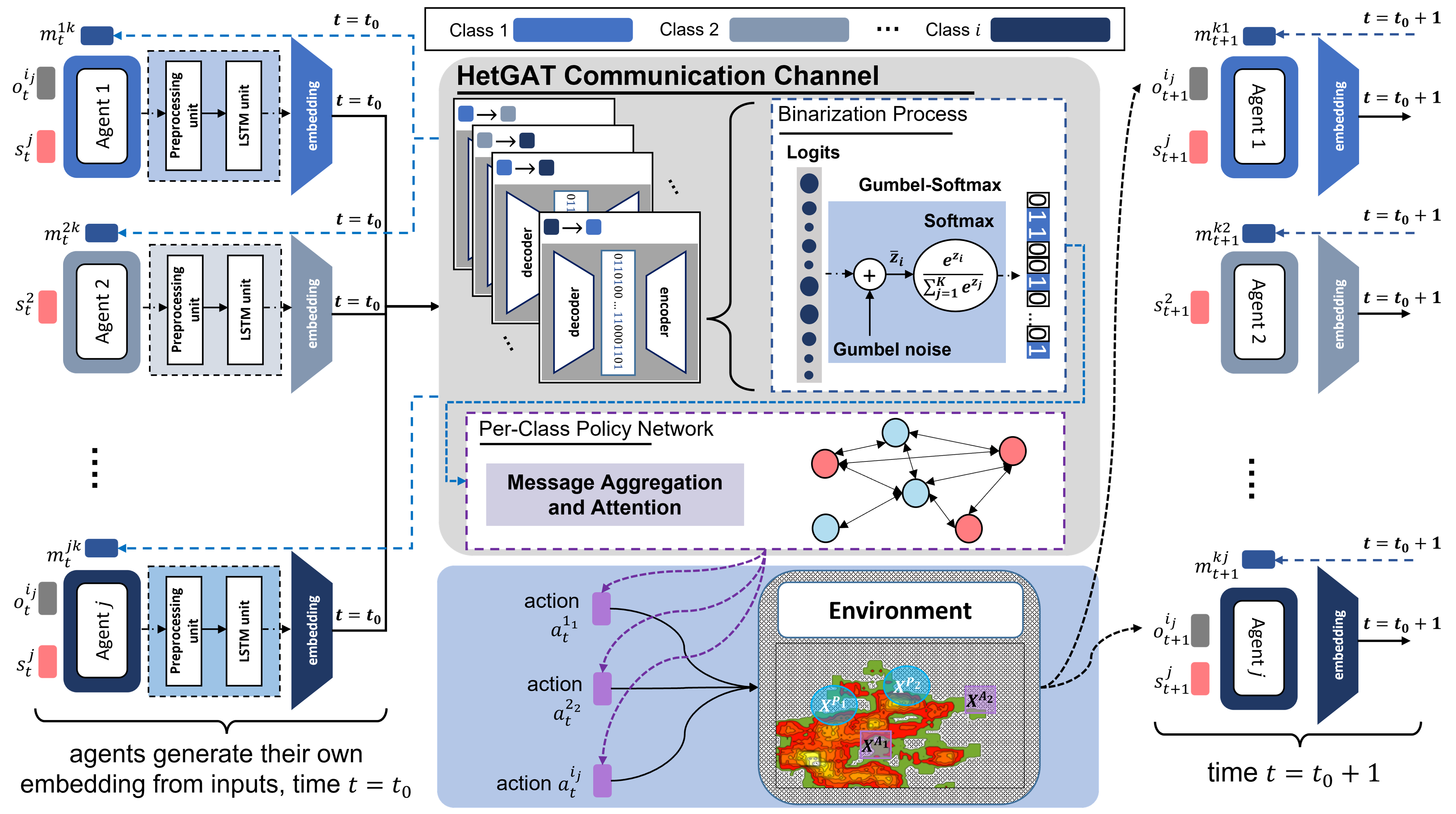}
	\caption{Overview of our multi-agent heterogeneous attentional communication architecture \textcolor{black}{in a CTDE paradigm}. At each time point $t=t_0$, each agent $j$ of class $i$ generates a local embedding from its own inputs, by passing its input data through class-specific preprocessing units and an LSTM cell. Each agent then sends the embedding to a class-specific encoder-decoder networks to generate a binarized message, $m_t^{jk}$, from its local neighbor $k$. The message information is then decoded and aggregated via learned attention coefficients by the receiving agent to compute the action probabilities as its policy output.}
	\label{fig:FullMessagePassingSystem}
	\vspace*{-0.25cm}
\end{figure*}

\subsection{Heterogeneous Communication Model}
\label{subsec:HeterogeneousGraphForCompositeMulti-AgentTeams}
\noindent GNNs previously used in MARL operate on homogeneous graphs to learn a universal feature update and communication scheme for all \textcolor{black}{agents~\cite{Jiang2020GraphCR, sheng2020learning, liu2019multi, niu2021multi},} which fail to explicitly model the heterogeneity among agents. We instead cast the cooperative MARL problem into a heterogeneous graph structure, and propose a novel heterogeneous graph-attention network capable of learning diverse communication strategies based on agent classes. Compared to homogeneous graphs, a heterogeneous graph can have nodes and edges of different types that can have different types of attributes. This advantage greatly increases a graph's expressivity and enables straightforward modeling of complicated, composite teams.

Given our MAH-POMDP formulation in the Preliminaries Section, we directly model each agent class $i\in\mathcal{C}$ as a unique node type.
This \textcolor{black}{approach} allows agents to have different types of state-space content, $\mathcal{S}^{(i)}$, as input features according to their classes, as well as enabling different types of action spaces, $\mathcal{A}^{(i)}$. Communication channels between agents are modeled as directed edges connecting the corresponding agent nodes. When two agents move to a close proximity of each other such that \textcolor{black}{those agents} fall within communication range, we add bidirectional edges to allow message passing between them. We use different edge types to model different class combinations of the sender and receiver agents to allow for learning heterogeneous communication protocols and intermediate representations.

Accordingly, we present an overview of our multi-agent heterogeneous attentional communication architecture in \textcolor{black}{Figure}~\ref{fig:FullMessagePassingSystem}. At each time, $t$, the features of each agent (i.e., each node of the heterogeneous graph) are generated through a class-specific feature preprocessor. We utilize separate modules to preprocess an agent's state, $s_t^{i_j}$, and observation, $o_t^{i_j}$, since depending on the agent's class, the environment observation input may not be available. 
\textcolor{black}{Each preprocessing module contains a Fully-Convolutional Network (FCN) \cite{Shelhamer2014FullyCN} Unit followed by an LSTM Unit to enable reasoning about both spatial and temporal information, further discussed in the Preprocessing Tensorized Representations Section \ref{subsec:tensor_representation}} and visualized in \textcolor{black}{Figure}~\ref{fig:rfcn_layers_vis}. As shown in \textcolor{black}{Figure}~\ref{fig:FullMessagePassingSystem}, the generated embeddings are then passed into a HetGAT communication channel, including a class-specific encoder-decoder network and a Gumbel-Softmax~\cite{jang2016categorical} unit to generate a binarized message, $m_t$, for an agent, $j$. 

\subsection{\textcolor{black}{Preprocessing Tensorized Representations}}
\label{subsec:tensor_representation}
\textcolor{black}{
\textcolor{black}{In this evolved version of HetNet, we are concerned with scaling heterogeneous robot teams to multi-agent domains with larger environment sizes and larger team sizes, varying in its compositionality, to allow for both effective learning within a single configuration and transferring knowledge to larger configurations that require significantly more exploration.} Prior works have designed communicative MARL frameworks with input feature representations, namely \emph{vectorized} inputs, using fixed-length one-hot encodings to encode an agent's state and observation. These representations are specific to a domain configuration (\textcolor{black}{e.g., }environment size and observation range) and do not effectively capture spatial information. As such, the learned policies designed for vectorized inputs cannot be deployed to domain configurations with larger environments and larger team sizes, as this would require additional parametrization to handle increased input dimensions. This inability to be non-parametric to task configuration size disqualifies prior work from utilizing transfer learning methods to warm-start learning in large-scale domain configurations, where learning can be much more difficult and computationally expensive.}

\begin{figure}[b!]
    \centering
  \includegraphics[width=1\columnwidth]{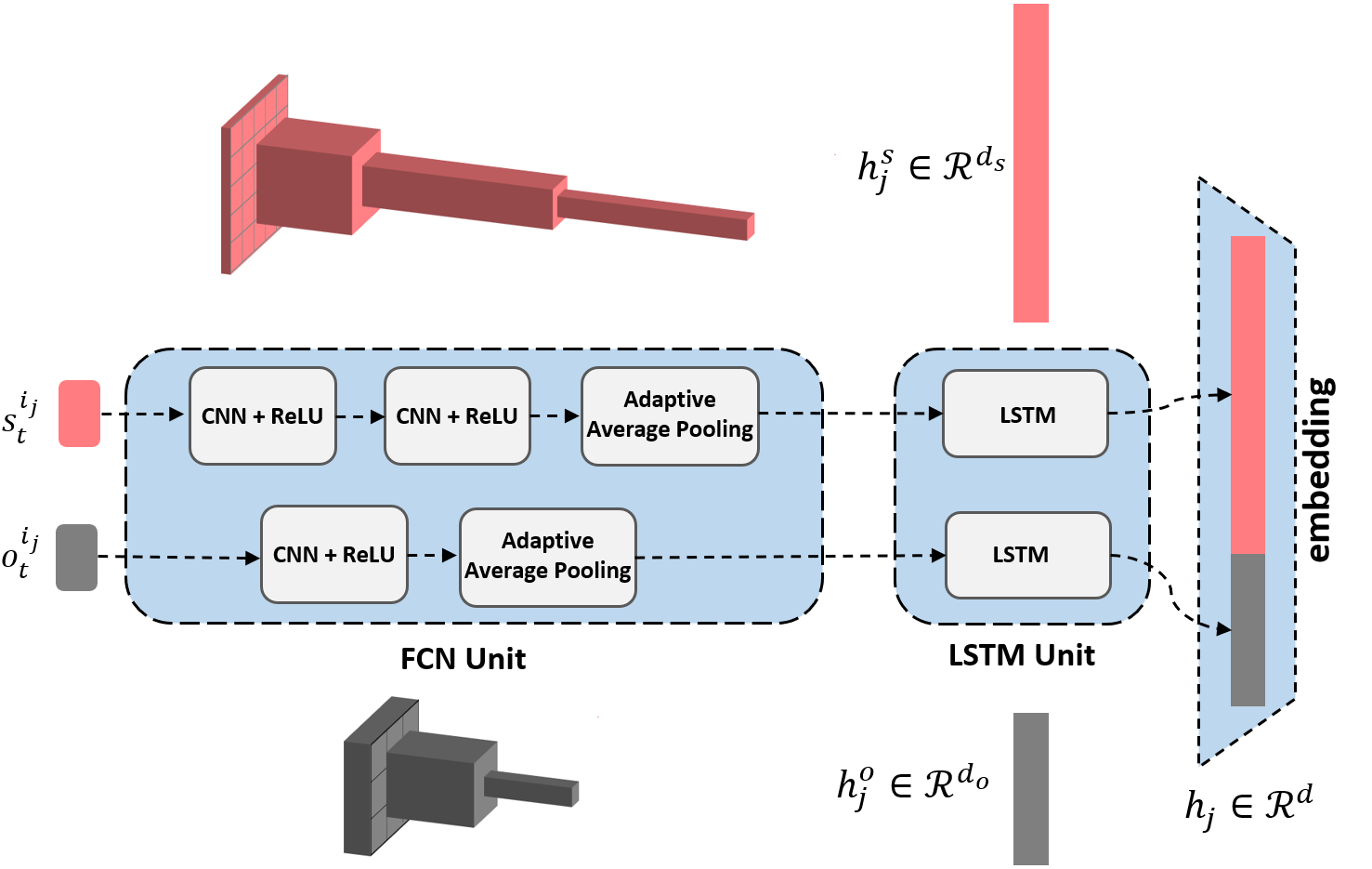}
  \caption{\textcolor{black}{Visualization of our proposed preprocessing architecture. Separate FCN Units, for state and observation inputs, generate fixed-length embeddings from size-varying tensorized input representations. The hidden embeddings of the following LSTM units are combined, generating a fixed-length embedding for the HetGAT communication channels.}}
  \label{fig:rfcn_layers_vis}
  \vspace{-5mm}
\end{figure}

\begin{figure}[t!]
    \centering
  \includegraphics[width=1\columnwidth]{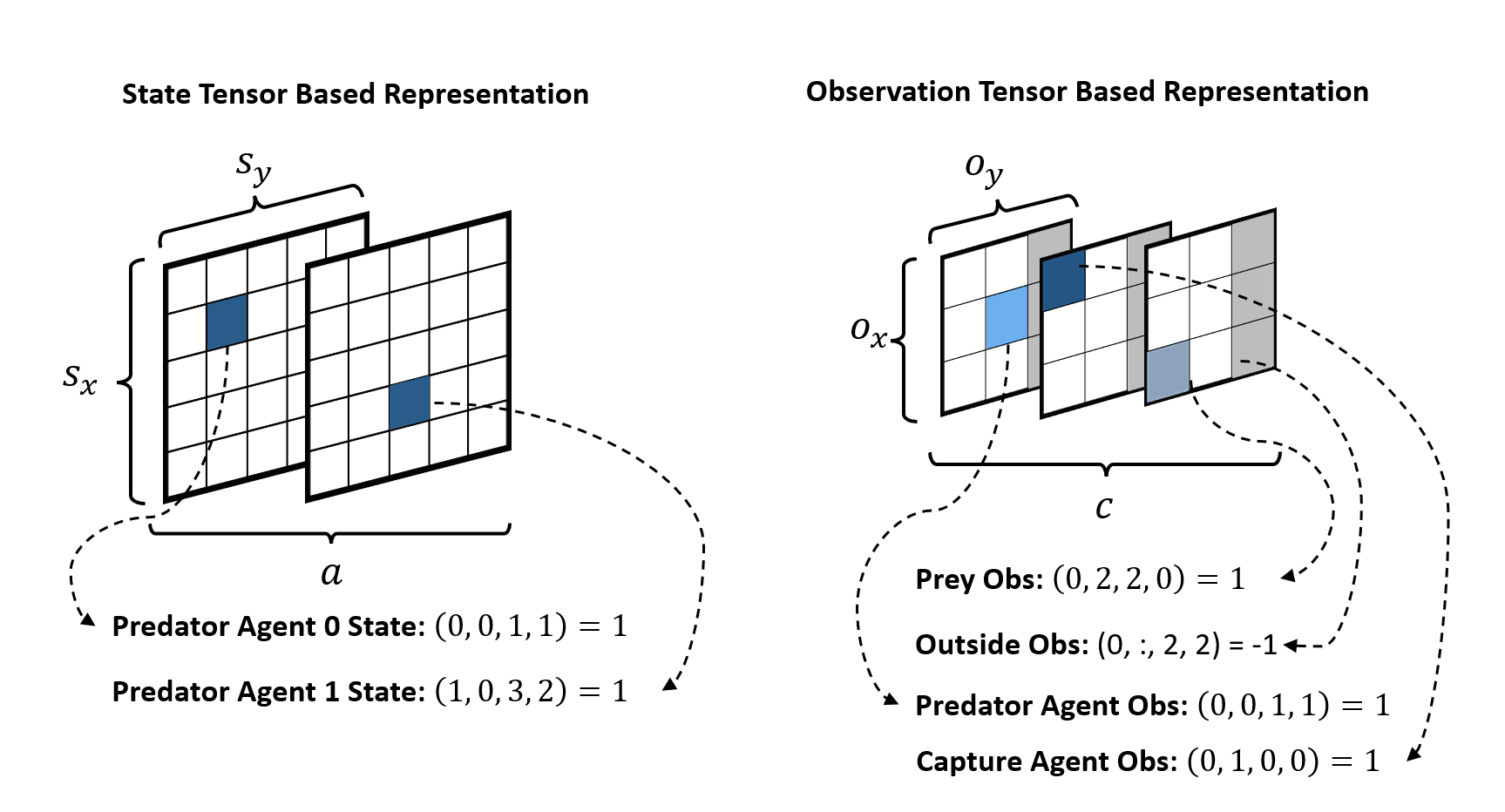}
  \caption{\textcolor{black}{Proposed tensorized input representations for an agent's state and observation, utilizing multi-dimensional tensors with dimensions $\big(i, c, s_{x}, s_{y}\big)$ and $\big(i, c, o_{x}, o_{y}\big)$. 
  The visualization shows the encoding for a $5 \times 5$ PCP environment with a $3 \times 3$ observation range.}}
  \label{fig:tensorized_representation}
  \vspace{-5mm}
\end{figure}

\emph{We improve the existing HetNet policy architecture to support utilizing a variable-sized, \emph{tensorized} feature representation of an agent's state and observations.}
This enables us to transfer coordination policies learned in smaller domain configurations in order to warm-start learning in larger, more challenging domain configurations. We define the state and observation tensorized representations with the multi-dimensional tensors $\big(i, c, s_{x}, s_{y}\big)$, and $\big(i, c, o_{x}, o_{y}\big)$, respectively. Across both tensors, the $i$ dimension indexes an agent's state or observation in the environment. The cardinality of this dimension is determined by the number of agents in a particular class, as agents may not have observations in our heterogeneous setting. Furthermore, the channel dimension, $c$, corresponds to an agent's class. For the state tensor, the cardinality of this dimension is set to one, as only one channel is necessary to encode the state. For the observation tensor, the cardinality of this dimension is determined by the number of classes in the heterogeneous domain. The last two dimensions encode spatial information about the state or observation of the \textcolor{black}{robot} agent, namely positions in the environment. For the state tensor, the cardinality of the $s_{x}$ and $s_{y}$ dimensions correspond to the size of the environment. For the observation tensor, the cardinality of the $o_{x}$ and $o_{y}$ dimensions correspond to an agent's observation range, and positional encodings are relative to the agent's reference frame. As a brief example, as depicted in \textcolor{black}{Figure} \ref{fig:tensorized_representation}, the state representation of the location of Agent 1 of Class 0, at grid location $(3, 2)$ in a $5 \times 5$ environment, would have the following tensor value: $(1, 0, 3, 2) = 1$.

\textcolor{black}{We design preprocessing modules that can process tensorized input representations of varying dimensions in the size of the environment and the number of agents in the team. The architecture of a preprocessing module, shown in \textcolor{black}{Figure}~\ref{fig:rfcn_layers_vis}, consists of an FCN unit to reason about spatial information and an LSTM unit to reason about temporal information. Tensorized state and observation inputs are processed by separate preprocessing modules, as an agent may or may not have input observations, as detailed in our MAH-POMDP formulation in Section~\ref{subsec:ProblemFormulation}. The output of the LSTM Unit is denoted by $h^{s}_{j} \in \mathbb{R}^{d_s}$ and $h^{o}_{j} \in \mathbb{R}^{d_o}$, for each preprocessing module, where $d_s$ and $d_o$ are the feature dimensions of state and observation embeddings, respectively. These embeddings are concatenated such that the input feature embedding to the HetGAT communication channel is denoted as $h_{j} = [h^{s}_{j} \Vert h^{o}_{j}] \in \mathbb{R}^{d}$, where $d = d_s + d_o$.}

The purpose of our FCN architecture is twofold: 1) the Convolutional Neural Network (CNN) layers are able to better encode spatial and class-type information by processing multiple, grid-like channels, and 2) the Adaptive Average Pooling layer is able to generate a fixed-length embedding, regardless of varying size tensorized representation inputs. Together, these properties enable us to learn better policies in large-scale heterogeneous domains and enable transferring a learned policy to new domain configurations without the need to add additional parametrization as required by prior work. 

\textcolor{black}{Several works in MARL \cite{balachandar2019collaboration, han2019grid, schwab2020tensor, pimentel2022scaling} utilize tensor-based representations for representing multiple agents within a grid, as these representations can be utilized across tasks. The closest related work to ours is \cite{schwab2020tensor}, which also provides results transferring a policy trained on a small domain with few agents to a larger domain with more agents. However, their MARL formulation uses a \textcolor{black}{fully-observable} MDP, while ours is a POMDP augmented with message communication between agents \textcolor{black}{to mitigate partial observability}. We note while this formulation limits our approach to domains where state and observation information can be represented as tensors, this space represents a variety of domains, including applications ranging from adversarial team versus team battles within the StarCraft Multi-Agent Challenge \cite{Samvelyan2019TheSM} to drone-fleet navigation for radiological inspection \cite{vale2023heterogeneous}, where each drone utilizes voxel-based representations.}

\subsection{Binarized Communication Channels}
\label{subsec:HetGATwithDigitizedCommunicationChannel}
\begin{figure}[t!]
	\centering
	\includegraphics[height=9.5cm, width=\linewidth]{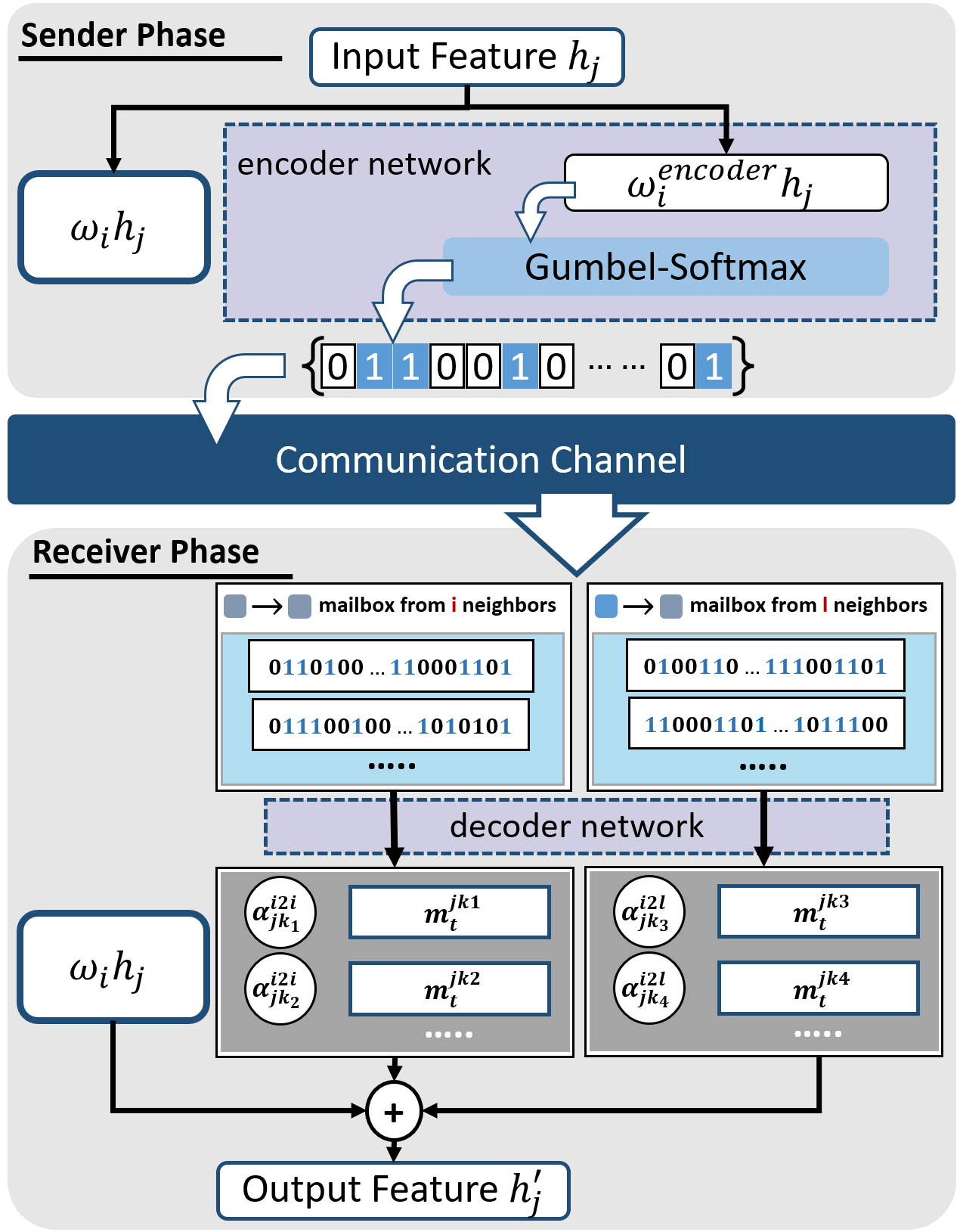}
	\caption{The sender and receiver phases of the feature update process in a HetGAT layer for \textcolor{black}{one} agent, $j$, of class $i$.}
	\label{fig:graphcompute}
	\vspace*{-0.5cm}
\end{figure}

\noindent The feature update process in a HetGAT layer is conducted in two steps: \textit{per-edge-type} message passing followed by \textit{per-node-type} feature reduction. When modeling multi-agent teams, we reformulate the computation process into two phases: a \textit{sender} phase and a \textit{receiver} phase. \textcolor{black}{Figure}~\ref{fig:graphcompute} shows the computation flow during the sender and receiver phases for an agent, $j$, of class $i$.

During the sender phase, the agent, $j$, of class $i\in\mathcal{C}$, processes its input feature, $h_j$, using a class-specific weight matrix, $\omega_{i}\in\mathbb{R}^{d^\prime \times d}$, where $d$ and $d^\prime$ are the input and output feature dimensions, respectively. The agent also transforms $h_j$ into the assigned message dimension using a class-specific encoder, $\omega_{i}^{enc}\in\mathbb{R}^{n\times d}$, where $n$ is the communication channel bandwidth. Next, we leverage a universal binarization process utilizing Gumbel-Softmax to convert the message into $0$s and $1$s for all classes as an efficient, intermediate language. The binarized message is then sent to neighboring agents. 
\textcolor{black}{We note that, a class-specific encoder entails a per-edge-type communication mechanism, meaning that the communication encoders learn to encode the embedding information according to the classes of both the sender and the receiver agents. The meaning of this class-specific encoded message is then determined at the class-specific decoder, again based on the receiver agent’s class. Subsequently, the message decoded by each agent is directly integrated as part of the input space, along with the agent's state observation, to update the policy.}

During the receiver phase, \textcolor{black}{robot} agent, $j$, of class $i$, processes all the received messages using a class-specific decoder, $\omega^{dec}_i\in\mathbb{R}^{d^\prime\times n}$. Next, for each type of communication edge that an agent is connected to, the HetGAT layer computes per-edge-type aggregation result by weighing received messages, along the same edge-type with normalized attention coefficients, $\alpha^{edgeType}$. The aggregation results are then merged with the agent's own transformed embedding, $\omega_ih_j$, to compute the output features. The feature update formula for an agent is shown in \textcolor{black}{Equation}~\ref{eq:Pnode}, where $j$ and $k$ are agent indexes and, $i, l\in\mathcal{C}$ are class indexes; such that, $i2l$ is an $edgeType$ and means ``from class $i$ to class $l$''. $m_t^{jk}$ is the decoded message computed by \textcolor{black}{Equation}~\ref{eq:message1} and, $\Delta_l(j)$ include agent $j$'s neighbors that belong to class $l$.

\begin{equation}
    \label{eq:Pnode}
    \text{Class}~(i):~~~\bar{h}_j^\prime = \sigma\Big(\omega_i \bar{h}_j +\sum_{l\in\mathcal{C}}\sum_{k \in {N_l(j)}}\alpha_{jk}^{i2l} m_t^{jk}\Big)
\end{equation}
\begin{equation}
\label{eq:message1}
    m_t^{jk} = \omega^{dec}_i (\mathrm{GumbelSoftmax} ( \omega^{enc}_lh_{k} )  )
\end{equation}Note that we have $l=i$ for intra-class communication. When computing attention coefficients in a heterogeneous graph, we adapt \textcolor{black}{Equation}~\ref{eq:attention} into \textcolor{black}{Equation}~\ref{eq:attNew} to account for heterogeneous channels.
\begin{equation}
\label{eq:attNew}
    \alpha_{jk}^{i2l} = \mathrm{softmax}_k\left(\sigma^\prime\left(\bar{W}_{att}^T\left[\omega_i\bar{h}_j \mathbin\Vert \omega_{i2l} \bar{h}_k \right]\right)\right)
\end{equation}

To stabilize the learning process, \textcolor{black}{we adapt the multi-head extension of the attention mechanism~\cite{velickovic2018graph} to fit our heterogeneous setting.} We use $L$ independent HetGAT (sub-)layers to compute node features in parallel and then merge the results by concatenation operation for each multi-head sub-layer in HetNet except for the last layer, which employs averaging. As a result, each type of communication channel is split into $L$ independent, parallel sub-channels.

\subsection{Heterogeneous Policy Network (HetNet)}
\label{subsec:HetPolicyNetwork}
\noindent At each timestep, $t$, a HetGAT layer corresponds to one round of message exchange between neighboring agents and feature update within each agent. By stacking several HetGAT layers, we construct \textcolor{black}{the} Heterogeneous Policy Network (HetNet) model that utilizes multi-round communication to extract high-level embeddings of each agent for decision-making. For the last HetGAT layer in HetNet, we set each agent's output feature dimension to the size of its action-space, specific to its class, $i$. Then, for each agent node, we add a Softmax layer on top of its output to obtain a probability distribution over actions that can be used for action sampling, resulting in class-wise stochastic policies. Accordingly, the computation process of each agent's policy remains local for distributed execution.

\section{Training and Execution}
\label{sec:TrainingandExecution}
 Our prior work with Heterogeneous Policy Network \cite{Seraj2022LearningED} utilized a Multi-Agent Heterogeneous Actor-Critic (MAHAC) formulation to learn policies across several grid-world domains, varying in complexity, with a maximum dimensionality of 5x5. However, deploying Heterogeneous Policy Networks at scale requires 1) an effective learning algorithm that is sample efficient and able to learn in sparsely-rewarded environments, and 2) the ability to model realistic perturbations to communication signals. 

\subsection{\textcolor{black}{Multi-Agent Heterogeneous PPO (MAH-PPO)}}
\label{subsec:MAHAC}
 \textcolor{black}{Motivated by the pitfalls of MAHAC in scaling to larger domains, we sought to create a learning algorithm for heterogeneous settings that is more sample-efficient (reducing computation time) and performs well in multi-agent settings. As such, given the recent success of PPO in multi-agent settings, we extend MA-PPO \cite{yu2021surprising} and present our Multi-Agent Heterogeneous PPO (MAH-PPO) framework that directly accounts for heterogeneity in multi-agent scenarios by learning class-wise coordination policies. \textcolor{black}{This advancement allows us to learn high-performance policies under increased heterogeneity and size.} We assign one policy per existing class, $\pi^i\in\{\Pi\}^{\mathcal{C}}$, each of which is parameterized by $\theta^i$. We utilize a \emph{centralized critic} value function $V^{\phi}$, parameterized by $\phi$, which takes as input the global state, including the agents' joint-state $\bar{s_t}$, and environment states. The trained actor network on the heterogeneous graph contains one set of learnable weights per agent class, which can be distributed to individual agents in the execution phase. During training, a centralized critic with access to global state input is shared by all agents, regardless of class, and can be disabled during decentralized execution following standard practice in CTDE paradigms.}
 
\textcolor{black}{To create MAH-PPO, we modify PPO's Clipped Surrogate Objective (Equation \ref{eq:ppo_clip}) to account for heterogeneity in the multi-agent setting. In doing so, we obtain the per-class multi-agent Clipped Surrogate Objective in Equation (\ref{eq:mahppo_obj_clip}), where $\vert B \vert $ is the batch size of trajectories, $r_{\theta^{i}}^{j}$ is the per-class probability ratio of an agent $j$ with a policy of class $i$, $\hat{A}_{t}^{j}$ is the advantage estimate of agent $j$, and $\epsilon$ is a hyper-parameter defining the clip range.}
\textcolor{black}{
\begin{equation}
\footnotesize
    J^{\text{CLIP}}(\theta^i) = \frac{1}{\vert B \vert \mathcal{N}^{(i)}} \sum_{\tau \in B} \sum_{j=1}^{\mathcal{N}^{(i)}} \sum_{t=1}^{T} \text{min}\big( r_{\theta^{i}}^{j} \hat{A}_{t}^{j}, \text{clip}(r_{\theta^{i}}^{j}, 1-\epsilon, 1+\epsilon)\big) \label{eq:mahppo_obj_clip}
\end{equation}}
\normalsize
\noindent The per-class probability ratio in Equation (\ref{eq:mahppo_obj_ratio}) measures the probability ratio of class $i$'s policy, where $\bar{a}_t^{i_j}$ and $\bar{o}_t^{i_j}$ represent the joint actions taken and joint observations received (if applicable for class $i$) by agents at time, $t$. Additionally, $\bar{m}_t$ represents the message-vector received by agent $j$ from its neighbors.
\begin{equation}
    r_{\theta^{i}}^{j} = \frac{\pi_{\theta^{i}}\big(\bar{a}_{t}^{i_j} | \bar{o}_{t}^{i_j}, \bar{m}_{t}\big) }{\pi_{\theta^{i}_\text{OLD}}\big(\bar{a}_{t}^{i_j} | \bar{o}_{t}^{i_j}, \bar{m}_{t}\big)} \label{eq:mahppo_obj_ratio}
\end{equation}
The advantage estimates are computed via the GAE \cite{schulman2015high} method, where $\delta_{t}^{j}$ is agent $j$'s \textcolor{black}{Temporal Difference (TD)} residual with discount $\gamma$, and $V^{\phi}(\bar{s}_t)$ is the centralized critic, parameterized by $\phi$, with global state input $\bar{s}_t$.
\begin{equation}
    \hat{A}^{j}_{t} = \sum_{l=0}^{T}(\gamma\lambda)^{l}\delta^{j}_{t+l} \label{eq:mahppo_gae}
\end{equation}
\begin{equation}
    \delta_{t}^{j} = r_t + \gamma V^{\phi}(\bar{s}_{t+1}) - V^{\phi}(\bar{s}_t) \label{eq:mahppo_td}
\end{equation}
Additionally, we add a per-class entropy term, in Equation (\ref{eq:mahppo_obj}) to ensure exploration, where  $S$ is the policy entropy and $c_1$ is the entropy coefficient. 
\begin{equation}
    J^{\text{S}}(\theta^i) = c_1 \frac{1}{\vert B \vert \mathcal{N}^{(i)}} \sum_{\tau \in B } \sum_{j=1}^{\mathcal{N}^{(i)}} \sum_{t=1}^{T} S\big( \pi_{\theta^{i}}(\cdot | \bar{o}_{t}^{i_j}, \bar{m}_t) \big) \label{eq:mahppo_obj_s}
\end{equation}
Putting together Equations (\ref{eq:mahppo_obj_clip}-\ref{eq:mahppo_obj_s}), the policy for each class, $\pi^i$, is updated via gradient descent to maximize the MAH-PPO objective in Equation (\ref{eq:mahppo_obj}). 
\begin{equation}
    J(\theta^i) = J^{\text{CLIP}}(\theta^i) + J^{\text{S}}(\theta^i) \label{eq:mahppo_obj}
\end{equation}
Given the improvements in the learning ability of our MAH-PPO formulation, we are able to learn under a shared reward task setting where credit assignment is much more difficult compared to our prior work, \textcolor{black}{especially as we increase the complexity of the domain}. \textcolor{black}{Motivated to explore these challenges and following prior work \cite{yu2021surprising}, we train HetNet with shared rewards. However, we note that HetNet can technically be trained with individual and class-specific rewards under our MAH-POMDP formulation in Section \ref{subsec:ProblemFormulation}}. We provide further implementation details regarding our MAH-PPO implementation \textcolor{black}{and hyperparameters in Section II-B of the supplementary.}

\subsection{\textcolor{black}{Scalability via Transfer}} 
\label{subsec:transfer_method}
\textcolor{black}{Learning high-performance MARL coordination policies in large-scale domain configurations is difficult due to the increasing joint state-action space that agents must explore, resulting in an increasingly complex credit-assignment problem \cite{foerster2018counterfactual} and a large variance of policy gradients during training \cite{lowe2017multi}. Moreover, training MARL algorithms in large-scale domains has practical implications, including prolonged training times and the need for increased high-performance computational resources. In this section, we describe our approach for addressing these challenges by combining HetNet's ability to process varying-size tensorized input representations with transfer learning.}

\textcolor{black}{In our approach, we first train a HetNet policy on a simple domain configuration with a small environment size and a small number of agents. Training in this configuration ensures that we quickly achieve a high-performing \textcolor{black}{communication-coordination} policy with the ability to accomplish the task. We train a policy in this domain configuration until convergence before directly transferring to a domain configuration with a larger environment size and increased number of agents. We can transfer our policy in an end-to-end fashion, as our preprocessing modules accept size-varying tensorized inputs, without requiring additional parametrization. By directly transferring knowledge learned in small-scale domains, we are able to effectively and efficiently address the complexities of learning diverse large-scale communication protocols and intermediate language representations. This facilitates faster learning performance and enables policies to learn in large domain configurations where learning policies from scratch can fail.} 

\subsection{Lossy Communication}
\label{subsec:lossy_comm_formulation}
\color{black}
A ubiquitous assumption of multi-agent communication frameworks is the reliability of inter-agent communication. In practice, however, communication channels can be subject to signal perturbations from hardware failures, electromagnetic noise, and adversarial manipulation (\textcolor{black}{e.g., }signal jamming). While noise in digital communication has been explored extensively in works such as~\cite{Proakis2007, Otung2021Communication, Connelly1993Noise}, the study of noisy communication in MARL is scarce. In this section, we describe how we modify the communication channel of HetNet during training to account for noisy communication and introduce formulations for three different types of noisy channels. This artificial-manipulation communication scheme can be easily applied to other MARL communication frameworks to evaluate the performance of their learned policies when subject to imperfect communication channels.

\begin{figure}[t!]
    \centering
  \includegraphics[width=1\columnwidth]{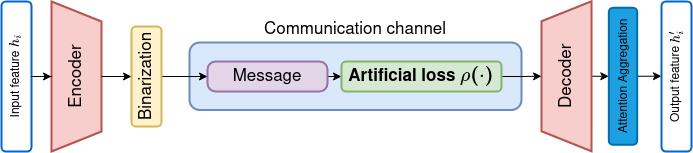}
  \caption{Visualization of modification to the communication channel for artificial loss}
  \label{fig:loss_high_level}
\end{figure}

To simulate such loss in our framework, we introduce artificial bit perturbation in the communication channel between agents. \textcolor{black}{Figure}~\ref{fig:loss_high_level} shows a high-level overview of where artificial loss is introduced in our communication pipeline. The input features of a sender agent $k$, $h_k$ are processed by its class encoder and binarized to produce a compressed and efficient message for transmission. Then, a form of loss is applied just prior to the decoding of the message by the receiving agent. By doing so, we can accurately represent where the majority of real-world noise is introduced, i.e. during the transmission of the message in the communication channel. The decoded message computed in \textcolor{black}{Equation}~\ref{eq:message1} is adapted into \textcolor{black}{Equation}~\ref{eq:lossy_message} to account for a noisy channel. 

\begin{equation}
\label{eq:lossy_message}
m_t^{jk}=\omega_i^{dec}(\rho(\text{GumbelSoftmax}(\omega_l^{enc}h_k)))
\end{equation}

\noindent In this equation, $\rho$ represents an arbitrary model of the communication channel that outputs a noisy message. We note that $\rho$ is non-differentiable due to the stochasticity of communication noise, thus prohibiting backpropagation during training. In our experiments, we evaluate three types of noisy communication channels over HetNet-Binary with tensor-based representation of agent observations. The first is a communication channel with static white Gaussian noise, where the noise level between all agents is fixed throughout a given experiment. This is the most basic scenario that allows us to measure the impact of noise as its intensity increases. The second noise model incorporates the distance between communicating agents, enabling an evaluation of our method's sensitivity to dynamically changing noise. The third model is a type-based noise model with different noise levels conditioned on the classes of the two communicating agents. Heterogeneous agents, by definition, have diverse traits, which include the electrical hardware for communication signals. Type-based noise allows us to capture noise that can result from heterogeneity and measure its impact on our architecture. In all three implementations, the noisy channel $\rho$ is parameterized by the bit error rate (BER) $p_b$ in \textcolor{black}{Equation}~\ref{eq:noise_function}.

\begin{equation}
\label{eq:noise_function}
\rho_{X\sim Bern(p_b)}(b_i)=
    \begin{cases}
        b_i & \text{if $X = 0$}\\
        1-b_i & \text{if $X = 1$}   
    \end{cases}, \forall b_i \in m_e
\end{equation}
In this equation, $m_e$ is the encoded message from the GumbelSoftmax that performs message binarization, as indicated in \textcolor{black}{Equation}~\ref{eq:lossy_message}.

\textbf{Additive White Gaussian Noise (AWGN)}: In an AWGN channel with binary phase-shift keying (BPSK) modulation, the BER $p_b$ is given in~\cite{Proakis2007} by: 

\begin{equation}
\label{eq:awgn_eq}
p_b = Q\left(\sqrt{\frac{2\varepsilon_b}{N_0}}\right)
\end{equation}

\begin{equation}
\label{eq:Q_function_eq}
Q(x) = P[\mathcal{N}(0,1)>x] = \frac{1}{\sqrt{2\pi}}\int_{x}^{\infty}e^{-\frac{t^2}{2}}dt
\end{equation}

$\frac{\varepsilon_b}{N_0}$ is the signal-to-noise (SNR) ratio per bit. The SNR is negatively correlated with the BER, i.e. decreasing the SNR increases the chances of a bit flip and vice versa.

\textbf{Range-based Noise with AWGN}: By the inverse square law, the intensity of a signal is inversely proportional to the square of the distance from the signal source. This property is expressed in the range-based noise model by scaling the signal strength with respect to the distance between two communicating agents. \textcolor{black}{Equation}~\ref{eq:awgn_eq} is adapted into \textcolor{black}{Equation}~\ref{eq:range_ber} to reflect this property.

\begin{equation}
\label{eq:range_ber}
p_b = Q\left(\sqrt{\frac{2\varepsilon_b}{N_0}\cdot\frac{1}{D^{jk}} }\right)
\end{equation}
In \textcolor{black}{Equation}~\ref{eq:range_ber}, $D^{jk}$ is the Euclidean distance between two agents $j$ and $k$ in our 2D-grid space environment. Each pair of agents is exposed to dynamic noise that changes with each time step.

\textbf{Type-based Noise with AWGN}: To account for the diverse types of transmitter, receiver, and modulation schemes in digital communication design, we introduce a general noise model to capture signal attenuation that can occur between different types of agents based on their communication scheme. \textcolor{black}{Equation}~\ref{eq:awgn_eq} is adapted into \textcolor{black}{Equation}~\ref{eq:type_ber} to model these deficiencies that can arise from the heterogeneity of the agents.

\begin{equation}
\label{eq:type_ber}
p_b = Q\left(\sqrt{\gamma^e\cdot\frac{2\varepsilon_b}{N_0}}\right), 0 \leq \gamma^e \leq 1
\end{equation}
In \textcolor{black}{Equation}~\ref{eq:type_ber} $\gamma^e$ is a tunable signal attenuation constant that is specific for each edge type in the communication framework. Note that all $\gamma^e = 1$ is equivalent to the base noise model with AWGN.

\section{Empirical Evaluation}
\label{sec:eval}
In this section, we discuss our evaluation environments, baselines that we compare our proposed formulation against, and results. We provide a full implementation of our approach \href{https://github.com/CORE-Robotics-Lab/HetNet/tree/hetnet-ppo}{https://github.com/CORE-Robotics-Lab/HetNet/tree/hetnet-ppo}.
\subsection{Evaluation Environments}

\begin{figure*}[ht!]
	\centering
	\includegraphics[width=\textwidth]{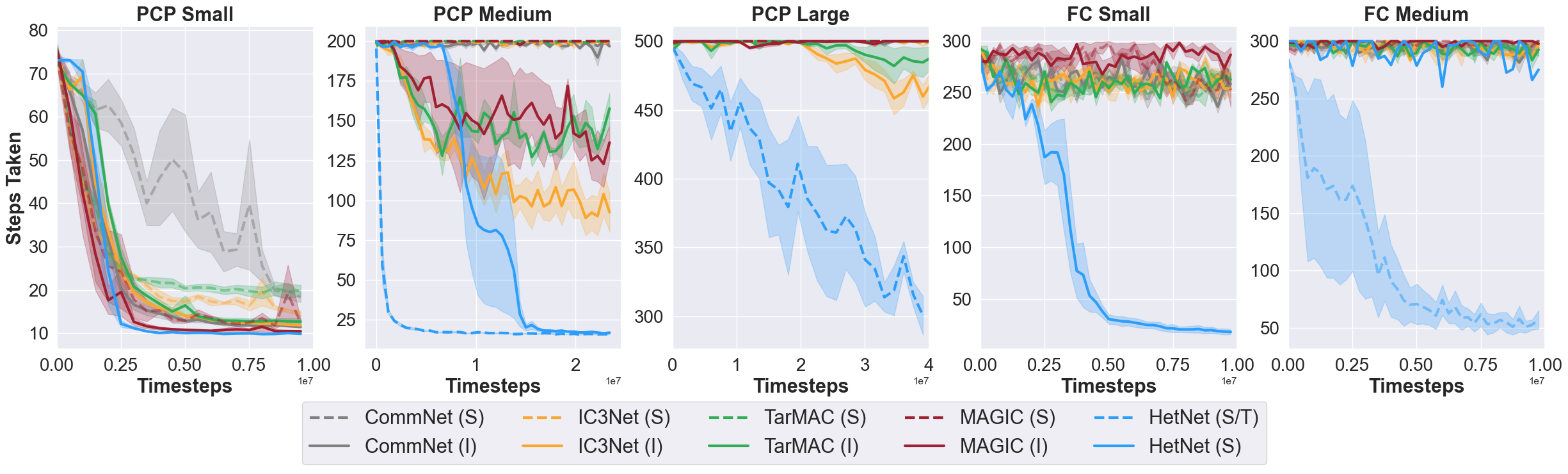}
	\caption{\textcolor{black}{
    Learning curves of training in the  PCP and FC domains. Shown is average steps taken ($\pm$ SE) by each method across episodes and three different random seeds as training proceeds. Baseline methods are trained with shared (S) and individual rewards (I). HetNet PPO (S) is trained with shared reward, and HetNet PPO (S/T) is trained with shared reward via transfer. HetNet outperforms all baselines in both domains in performance at convergence and sample efficiency.}
    }
	\label{fig:learning_curves}
	\vspace*{-0.25cm}
\end{figure*}

\label{subsec:evaluation_environments}
We evaluate the utility of HetNet against several baselines in two cooperative and heterogeneous MARL domains that require learning collaborative behaviors. These domains consist of two-class perception-action team of agents, where different classes of agents may have different observation and action spaces. \textcolor{black}{In this section, we describe the key objectives and the differences between the two domains. For complete descriptions and further details of the dynamics of each environment, please refer to Section I of the supplementary document.}

\textbf{Predator-Capture-Prey (PCP)~\cite{Seraj2022LearningED} --} For the first heterogeneous domain, we modify the homogeneous Predator-Prey \cite{singh2018learning} environment to create a new heterogeneous environment, which we refer to as Predator-Capture-Prey (PCP) \cite{Seraj2022LearningED}, to include a composite team.
In PCP, there are two classes of agents: \textit{predator} and \textit{capture} agents. The \textit{predator} class of agents have the ability to perceive the environment and have the goal of finding the prey with limited vision. We term these agents ``Perception" agents. The \textit{capture} class of agents have the goal of locating the prey \emph{and} capturing it with an additional \textit{capture-prey} action in their action-space, and importantly, do not have any observation inputs (e.g., lack of scanning sensors). We term these agents ``Action" agents. We define three configurations for this domain, varying the environment and team sizes, as in \cite{singh2018learning}. \textcolor{black}{The baseline configuration (\emph{small}) for the PCP environment is a size $5 \times 5$ grid-world with 2 Perception (P) and 1 Action (A) agents. In this configuration, the perception agent's vision is limited as they only observe a $1 \hspace{-0.5mm} \times \hspace{-0.5mm} 1$ grid around them. Likewise, we define the \emph{medium} configuration as a size $10 \times 10$ grid-world with 3P and 2A agents, and the \emph{large} configuration as a $20 \times 20$ grid-world with 6P and 4A agents. In these configurations, perception agents can observe a $3 \hspace{-0.5mm} \times \hspace{-0.5mm} 3$ grid around them.}

\textbf{FireCommander (FC)~\cite{seraj2020firecommander} --} In the second heterogeneous domain, the FireCommander~\cite{seraj2020firecommander}, two classes of \textit{perception} and \textit{action} agents must collaborate as a composite team to extinguish a propagating firespot. At each timestep, the firespot propagates to a new location according to the FARSITE~\cite{finney1998farsite} model, while the previous location is still on fire. All firespots are initially hidden to agents and need to be discovered before being extinguished. As such, \textit{perception} agents are tasked to scan the environment to detect the firespots while \textit{action} agents (no observation inputs) are required to move and extinguish a firespot that has been discovered by a \textit{perception} agent before. Note that since firespots propagate, both \textit{perception} and \textit{action} agents need to continue to explore the map and collaborate until all firespots are extinguished. \textcolor{black}{We define the domain configurations for the FC environment the same as in PCP, however, perception agents have increased observation range as they can observe a $3 \times 3$ grid around them in both \emph{small} and \emph{medium} configurations.}

\subsection{Baselines}
We benchmark our framework against four end-to-end communicative MARL baselines: (1) CommNet~\cite{sukhbaatar2016learning}, (2) IC3Net~\cite{singh2018learning}, (3) TarMAC~\cite{das2018tarmac} and, (4) MAGIC~\cite{niu2021multi}.
We note that, for all four baselines we directly pulled the respective authors' publicly available code-bases and hyperparameters for training \textcolor{black}{to obtain best-effort baseline implementations for comparison}. 
\textcolor{black}{Additionally, to further evaluate the effectiveness of MAH-PPO, we compare against HetNet trained with MAHAC as in prior work \cite{seraj2022learning}.}
All baseline methods are trained using a default vectorized input representation. 
For fair comparison, we modify each baseline's critic to accept the environment-provided global

\begin{figure*}[ht!]
	\centering
	\includegraphics[width=\textwidth]{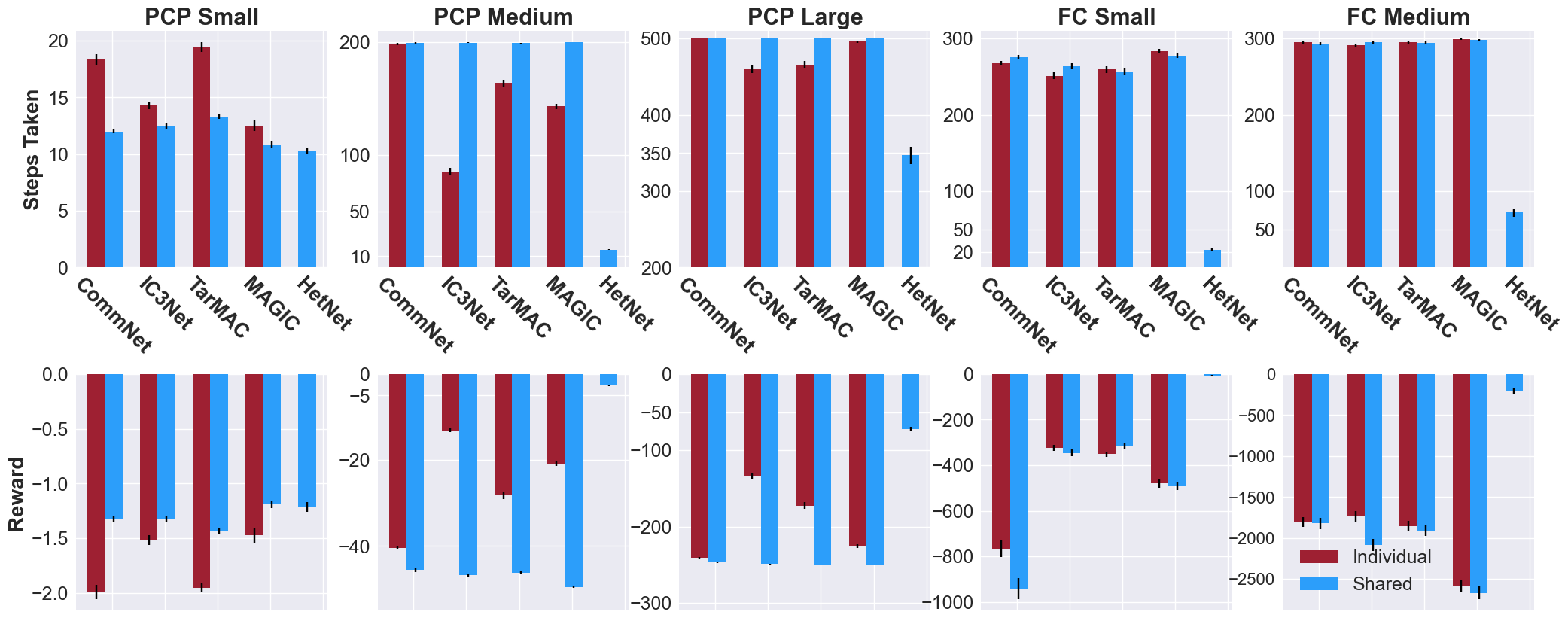}
	\caption{\textcolor{black}{Evaluation results across domain configurations}. Reported results are averaged (± standard error (SE)) from 100 evaluation trials with different random-seed initializations. For all tests, the final training policy at convergence is used for each method. As shown, HetNet outperforms all baselines in both heterogeneous domains.}
	\label{fig:evaluation_results}
	\vspace*{-0.25cm}
\end{figure*}

\subsection{Results and Discussion}
\noindent In this section, we empirically validate the performance of our frameworks across several heterogeneous teaming domains against the introduced baselines. Across each domain, we seek to minimize the number of steps taken to accomplish the search-and-rescue problem.

\subsubsection{Performance Comparison To Baselines}
\label{subsubsec:baseline_comparison_results}
We compare HetNet's performance against other baseline methods on the \emph{small} domain configurations of the PCP \cite{Seraj2022LearningED} and FC \cite{seraj2020firecommander} domains. \textcolor{black}{Figure}~\ref{fig:learning_curves} depicts the average steps taken ($\pm$ standard error) by each method across episodes as training proceeds in each configuration of both domains. In both domains, PCP and FC, HetNet outperforms all baselines, trained with either shared (S) or individual (I) reward, by converging to a more efficient coordination policy (i.e., fewer steps taken).

Furthermore, we evaluate the learned coordination policies at convergence by each of the baselines in both domains. \textcolor{black}{For fair comparison, we randomly generate a set of 100 initial conditions for the positions of agents and prey or fire locations, for each domain configuration, and evaluate each method with these same initial conditions.} The results of this evaluation are presented in \textcolor{black}{Figure}~\ref{fig:evaluation_results}, where the reported results are the average ($\pm$ standard error (SE)) steps taken from 100 evaluation trials. As shown, in both the PCP and FC domains, HetNet outperforms all baselines trained with either a shared or individual reward scheme. We highlight that only HetNet demonstrates the ability to learn a high-performing policy in the \emph{small} configuration of the FC domain, which is of higher complexity than the PCP domain. HetNet's superior performance can be explained by its ability to handle heterogeneity, while baseline methods struggle as they are not designed for this purpose. 

We also note that baseline methods perform better when trained with shared reward, compared to individual reward within the same configuration. This can be attributed to shared reward enabling agents to learn policies that achieve the composite $team's$ goal rather than learning policies that benefit the skill of a particular class of agent. Our results indicate that this can be particularly important for heterogeneous teaming, as training with individual rewards can lead to learning sub-optimal policies. 

\textcolor{black}{Across both domains, HetNet achieves a 5.91\% to 1131.00\% performance improvement over other baselines, trained with either share or individual rewards, in the \emph{small} domain configurations.} The heterogeneous policies learned by our model set the SOTA for learning challenging cooperative behaviors for composite teams.

\begin{figure}[b!]
    \centering
    \includegraphics[width=\columnwidth]{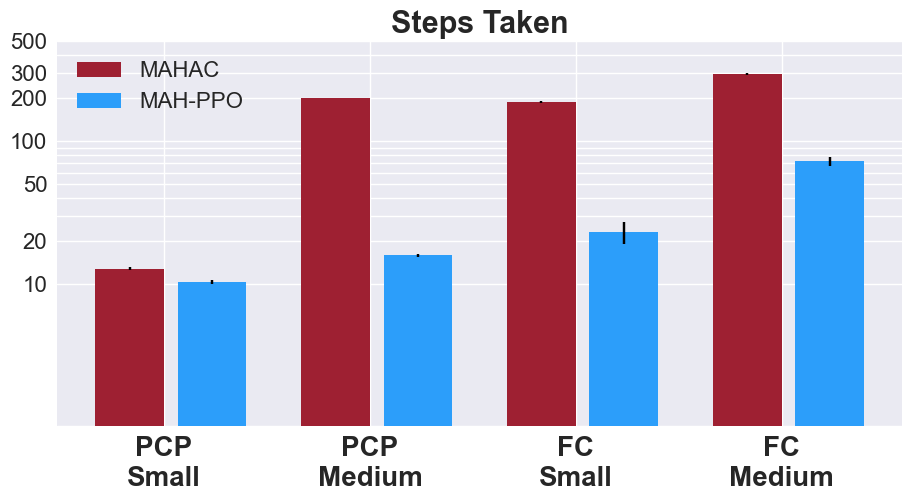}
    \caption{\textcolor{black}{Performance comparison between HetNet trained with MAHAC in prior work \cite{seraj2022learning} and HetNet trained with MAH-PPO. 
    As shown, HetNet trained with MAH-PPO outperforms HetNet trained with MAHAC across domain configurations.}}
    \label{fig:hetnet_a2c_vs_ppo}
\end{figure}

\subsubsection{\textcolor{black}{Performance Comparison to HetNet under MAHAC}}
\label{subsubsec:a2c_comparison_results} 
\textcolor{black}{In addition to other end-to-end communicative MARL baselines, we compare our method against our prior work that trains HetNet under MAHAC \cite{seraj2022learning} in both the Small and Medium configurations of PCP and FC. For a fair comparison to HetNet with MAH-PPO, we train HetNet with MAHAC under a shared reward scheme. We evaluate the trained policies under the same evaluation conditions described in Section \ref{subsubsec:baseline_comparison_results} and highlight our results in Figure ~\ref{fig:hetnet_a2c_vs_ppo}. We observe that MAH-PPO outperforms MAHAC in both domains and across both configuration scales. In the Small domain configuration, we observe that MAH-PPO achieves a $23.83\%$ to $1166.10\%$ performance improvement over MAHAC. Most notably, we observe that MAH-PPO is most beneficial in the more complex Medium domain configurations, as it achieves a $305.13\%$ to $1166.10\%$ performance improvement over MAHAC. Our results highlight the importance of designing advanced optimization methods, such as MAH-PPO, to enable learning in complex heterogenous domains where prior formulations fail to learn high-performing coordination policies. 
}
\textcolor{black}{
We run MAHAC using its public codebase release and corresponding settings to ensure implementation consistency with the other baselines presented in this paper.
We note that the HetNet-MAHAC convergence levels observed here do not fully match those reported in prior work \cite{seraj2022learning}. This discrepancy is not explained by the environment updates in this release and likely reflects implementation details or experimental conditions not fully specified in the earlier work. However, the overall relative comparison remains consistent.
}

\textcolor{black}{To supplement our empirical comparison to MAHAC, in Section II-B of the supplementary, we highlight several implementation improvements in training HetNet under MAH-PPO that enable feasible training in large-scale domains, such as PCP Large. In Section II-C of the supplementary, we compare the runtime performance of training HetNet with MAH-PPO and training with the implementation of MAHAC. We find that due to elongated training times, training HetNet with MAHAC in PCP Large is intractable. With the implementation improvements, HetNet trained with MAH-PPO leads to a 98.68\% decrease in experiment runtime.}

\subsubsection{Scalability}
\label{subsubsec:scalability_results}
\textcolor{black}{We conduct experiments to evaluate the scalability of our framework, against other baselines, to larger environments and team sizes in the \emph{medium} and \emph{large} configurations of the PCP and FC domain. In \textcolor{black}{Figure} \ref{fig:learning_curves}, we observe that HetNet, trained from scratch, converges much quicker than other baseline methods trained on the \emph{medium} configuration of the PCP domain. \textcolor{black}{As shown in \textcolor{black}{Figure} \ref{fig:evaluation_results}, HetNet PPO achieves a 32.5\% to 1166.10\% performance improvement over other baselines, trained in the \emph{medium} and \emph{large} domain configurations, outperforming all other baselines trained with either shared (S) or individual (I) reward.} Furthermore, we observe that in the \emph{medium} and \emph{large} configurations, baseline methods achieve higher performance when trained using individual rewards instead of shared rewards, as opposed to their performance on the \emph{small} domain configurations. This might be attributed to individual reward enabling agents to learn policies that explore the larger scale domain better in a way that benefits their individual skills (\textcolor{black}{e.g., }perception agents first learn to find the prey and then learn how to communicate this information to action agents). Moreover, training with shared reward makes the credit-assignment problem more difficult \cite{foerster2018counterfactual}. Despite this challenge, HetNet, trained with shared reward, is able to overcome the difficulty in exploration and converge to a much higher-performing policy than other baselines, displaying the proficiency of the adapted architecture of HetNet and associated MAH-PPO formulation.}

\begin{figure*}[t!]
  \centering
  \begin{subfigure}{\columnwidth}
    \centering
    \includegraphics[width=\linewidth]{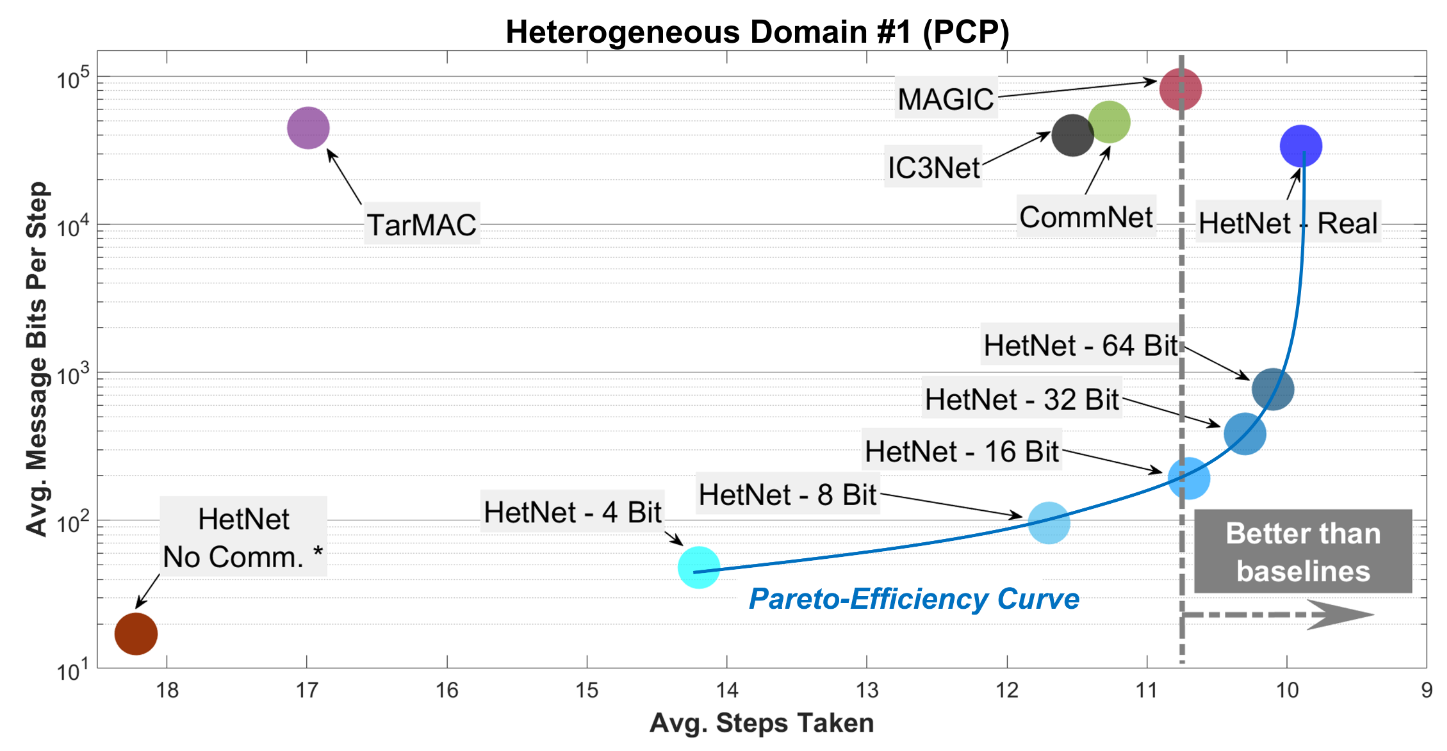}
    \caption{\textcolor{black}{CB vs performance results in in PCP domain as shown in~\cite{Seraj2022LearningED}}.}
    \label{fig:pareto_curve_old}
  \end{subfigure}
  \begin{subfigure}{\columnwidth}
    \centering
    \includegraphics[width=\linewidth]{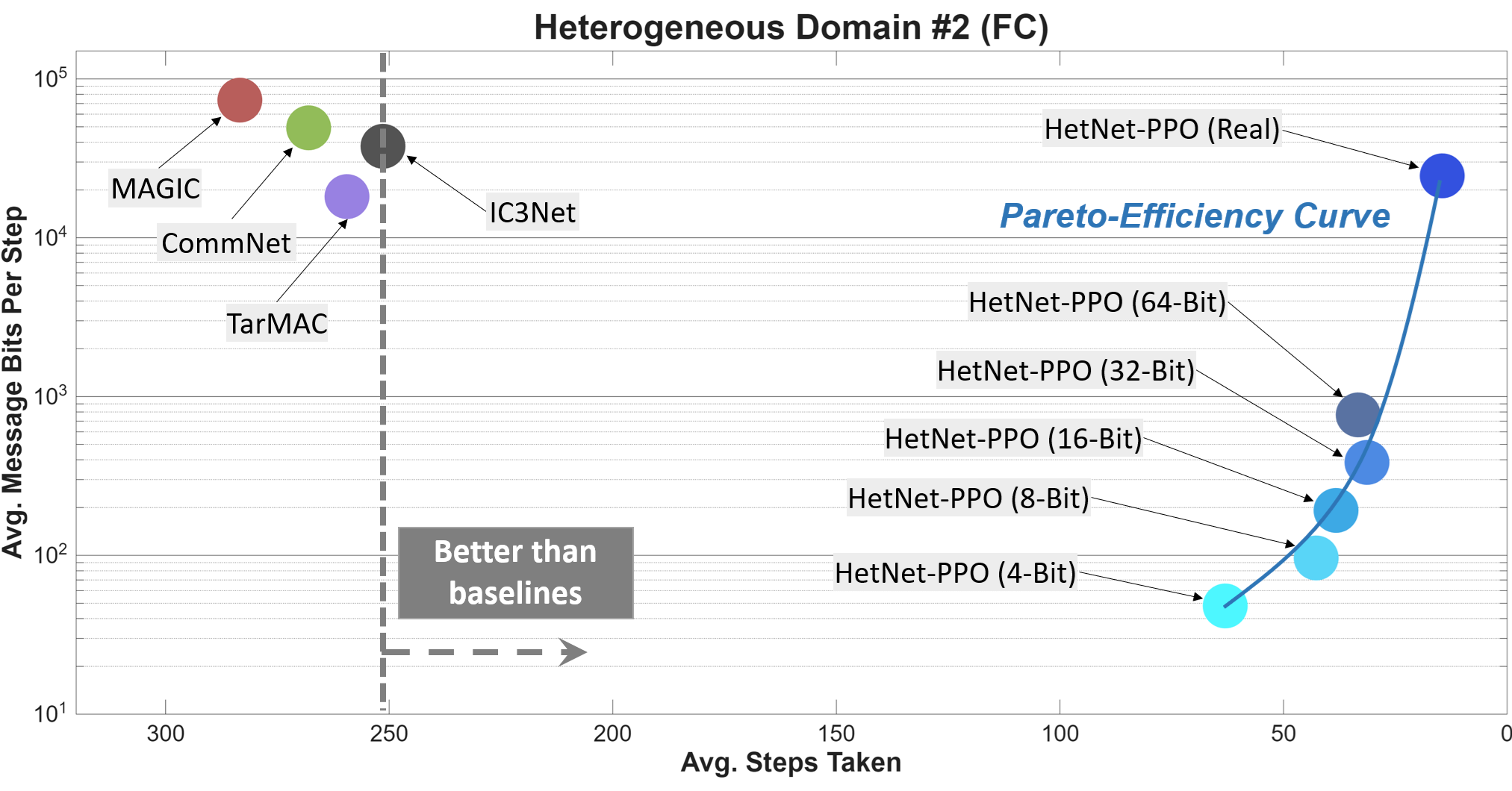}
    \caption{\textcolor{black}{CB vs performance results in the FC domain.}}
    \label{fig:fc_msg_dim_compare}
  \end{subfigure}
  \caption{\textcolor{black}{Results of experiments assesing Communication Bandwitdh (CB) as Communicated bits per round of communication. HetNet facilitates binarized messages among agents which requires significantly less CB as compared to real-valued baselines.}}
\end{figure*}

\begin{figure}[b!]
  \centering
  \begin{subfigure}{\columnwidth}
    \centering
    \includegraphics[width=0.8\linewidth]{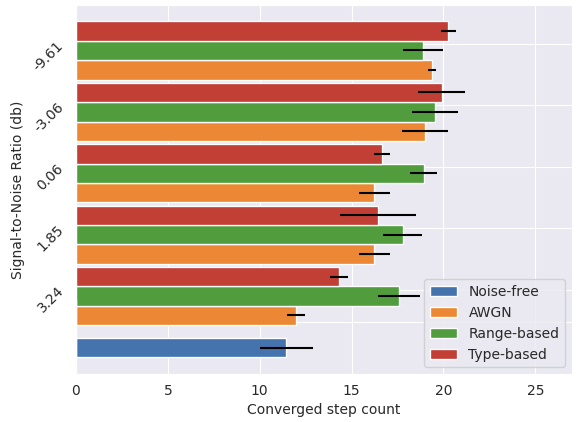}
    \caption{\textcolor{black}{Average steps taken across episodes at convergence under all noise conditions. All methods are trained with HetNet-PPO (S) and 16-bit messages.}}
    \label{fig:lossy_compare}
  \end{subfigure}
  \begin{subfigure}{\columnwidth}
    \centering
    \includegraphics[width=0.8\linewidth]{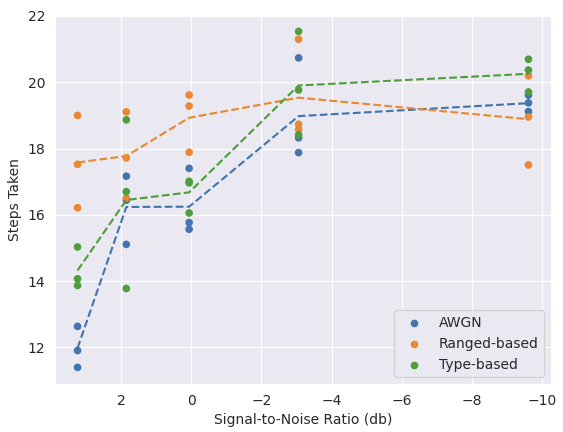}
    \caption{\textcolor{black}{Average steps taken across episodes as SNR decreases in \emph{small} domain configuration of PCP.}}
    \label{fig:lossy_scatter}
  \end{subfigure}
  \caption{\textcolor{black}{Results of experiments assessing the robustness of our architecture to communication noise. All results are averaged over three random seeds and experiments are performed in the \emph{small} domain configuration of PCP.}}
  \label{fig:two_subfigures}
\end{figure}

\subsubsection{\textcolor{black}{Transferability}}
\label{subsec:transferability_results}
\textcolor{black}{We show the utility of our proposed architecture, with the augmentations described in Section \ref{subsec:tensor_representation}, by scaling to both \emph{medium} and \emph{large} configurations of the PCP domain via transfer as described in Scalability via Transfer Section. \textcolor{black}{We pre-train a HetNet policy in the \emph{small} domain configuration with increased vision (\textcolor{black}{i.e., }an agent's observation range is $3 \times 3$) to ensure that we pre-train observation preprocessing modules for the same observation range as specified in the \emph{medium} domain configuration. We train this policy until convergence, transfer this learned policy to the more challenging \emph{medium} domain configuration with more agents and increased dimensionality, and continue training.} In \textcolor{black}{Figure}~\ref{fig:learning_curves}, we observe that the HetNet policy trained after transfer in the PCP \emph{medium} configuration is more sample efficient and converges much quicker than a policy trained from scratch. \textcolor{black}{Notably, it takes approximately 15 million training samples for the HetNet policy trained from scratch to achieve less than 20 steps per episode, while the transferred HetNet policy can achieve the same performance in approximately 3 million training samples. This represents a 389.51\% improvement in sample efficiency and an 80.63\% decrease in runtime required to achieve a high-performing policy.} We further leverage the transferability of our architecture and deploy a policy, trained until convergence in the \emph{medium} configuration, to the PCP \emph{large} configuration. In \textcolor{black}{Figure}~\ref{fig:learning_curves}, we observe that our transferred policy is able to achieve higher performance than other baselines trained from scratch in the PCP \emph{large} configuration.} Additionally, in \textcolor{black}{Figure}~\ref{fig:learning_curves} we observe the same benefits of transferability, as a HetNet policy trained to convergence in the FC \emph{small} configuration is transferred to the FC \emph{medium} configuration and is able to achieve higher performance than baselines trained from scratch. Our results demonstrate that our transferable architecture enables improved sample efficiency and enables learning in challenging domains where conventional training from scratch can fail.

\begin{figure*}[t!]
    \centering
    \includegraphics[width=\linewidth]{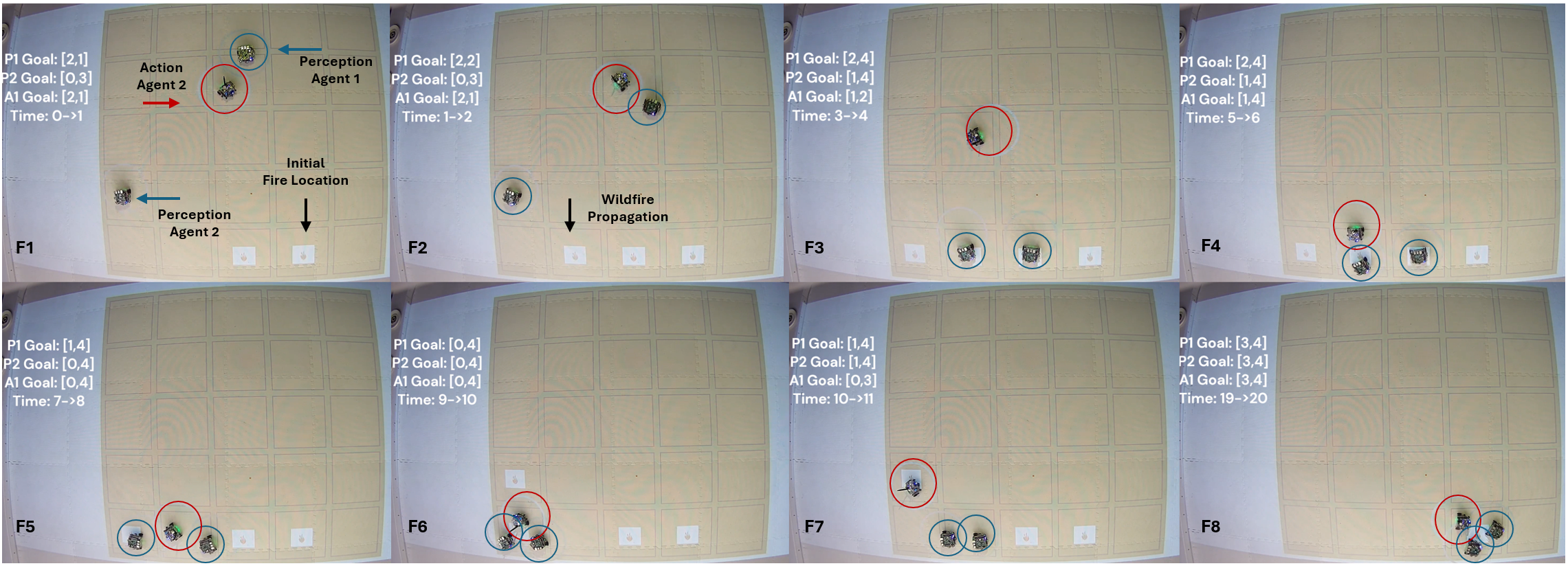}
    \caption{\textcolor{black}{Demonstration of our algorithm in the FireCommander domain on physical robots on the Robotarium platform. We highlight the different agents as well as the propagating wildfire through eight frames pulled from the demonstration video. In each frame, perception agents are circled in blue, while action agents are circled in red. We denote the goal location of each agent and the timestep of each frame on the top-right. The complete video and implementation details can be found in the supplementary material.}}
    \label{fig:robo}
\end{figure*}

\subsubsection{Robustness}
\label{subsec:lossy_comms_results}
\textcolor{black}{To assess the robustness of our architecture to communication noise, we perform a sensitivity analysis on HetNet when exposed to each of the three previously described noise models. We compare the performance of HetNet under the three noise models against HetNet with a noise-free channel to show the impact of noise on performance, as perfect communication serves as the lower bound on the number of steps that can be achieved. For all channel types, the network is trained with a tensorized input representation, shared reward, 16-bit binarized messages, and the same hyperparameters in the \emph{small} configuration of PCP. We chose a reduced message dimension to demonstrate the effectiveness of our architecture while under bandwidth constraints and noise simultaneously. For type-based noise, we fixed the signal attenuation constant $\gamma^e$ for edges between \emph{predator} and \emph{capture} agents to $0.8$ to model communication deficiencies between different classes of agents. The average steps taken at convergence for all three noise models ($\pm$ standard error) over a range of SNR is shown in \textcolor{black}{Figure}~\ref{fig:lossy_compare} and \textcolor{black}{Figure}~\ref{fig:lossy_scatter}. }

\textcolor{black}{Under the AWGN noise model, we observe a small performance decrease when the SNR is high. The performance continues to decay exponentially as SNR decreases and plateaus as the noise power overwhelms the signal power (\textcolor{black}{i.e., }when the SNR is negative). The final performance with negative SNR approaches the performance of HetNet with no communication ($\approx$20 steps) as shown in~\cite{Seraj2022LearningED}. With the range-based noise model, we observe a significant decrease in performance relative to the noise-free communication, even when the SNR is high. This indicates that our architecture is sensitive to high variability in noise during learning as the noise level changes based on the distance between agents. However, we note that the range-based noise condition with a positive SNR still outperforms scenarios with no communication. For type-based noise, we see that performance is similar to the AWGN condition with a slight decrease. This shows that our architecture is able to learn under noise introduced from the heterogeneity of composite teams. }

\textcolor{black}{In \textcolor{black}{Figure}~\ref{fig:lossy_scatter}, we additionally show the performance of our architecture with noise across all seeds and a range of SNR. We observe more variance in the final converged performance as opposed to HetNet (S) in \textcolor{black}{Figure}~\ref{fig:learning_curves}, which can be attributed to the binarization of messages, as similar variance can be observed with noise-free communication. In general, a higher SNR results in a decrease in the lower bound for average steps taken.}

\subsubsection{\textcolor{black}{Communication Bandwidth}}
\label{subsec:bandwidth_fc_results}
\textcolor{black}{We compute the Communication Bandwidth (CB), defined as the number of bits communicated per round, for HetNet trained under MAH-PPO (HetNet-PPO) across a range of message dimensionalities, and report performance in the \emph{small} domain configuration of FC in \textcolor{black}{Figure}~\ref{fig:fc_msg_dim_compare}.} The performance over CB follows a similar pareto-efficiency curve as shown in \textcolor{black}{Figure}~\ref{fig:pareto_curve_old} from our prior work~\cite{Seraj2022LearningED}. None of the baselines were able to converge in the FC domain. All binarized HetNet-PPO configurations were able to achieve better performance than the baselines, demonstrating the effectiveness of our framework under bandwidth restrictions.

\vspace{-0.3cm}
\color{black}{
\section{Real-World Robot Demonstration}
\label{sec:robo}
We present a demonstration of our algorithm emulating the \textit{Small} FireCommander example with two perception agents and one action agent on the Robotarium platform, a remotely accessible swarm robotics research platform \cite{Wilson2020TheRG}. In Figure \ref{fig:robo}, \textcolor{black}{we display several frames (F1-F8) of a trajectory being executed in this scenario} from a top-down perspective (not ego-centric to any agent), \textcolor{black}{where perception agents are encircled in blue and the action agent is encircled in red}. This demonstration showcases the feasibility of trajectories produced via HetNet on real robots and \textcolor{black}{the learned policy's coordination strategy throughout the complex FC scenario. Notably, we observe that in frame 2 (F2), perception agent 2 (PA2) becomes aware of a fire location as the fire propagates in the environment. We observe that this information is successfully communicated to action agent 1 (AA1), which cannot reason about fire locations due to its lack of sensory observations, and perception agent 1 (PA1). By F4, we observe all agents have moved to the fire locations and can observe AA1 putting out fires in later frames, F5 and F6. In F7, we highlight the coordination policy's strategy as AA1 navigates to the previously discovered propagating fire location, and both perception agents explore new locations for fire spots. At F8, all agents converge at the last fire location before AA1 puts out the last fire. Once this action is taken, the episode terminates.} We attach the full video of this demonstration in the supplementary material with annotations. \textcolor{black}{Furthermore, we include additional implementation details of this demonstration in Section III of the supplementary document.} We leave training MARL algorithms on real-world heterogeneous robots, which can require a large number of samples, to future work.}

\vspace{-0.5cm}
\section{Conclusion}
In this extended work, we create a scalable MARL architecture and associated learning algorithm to support heterogeneous robot team coordination, \textcolor{black}{advancing the state-of-the-art of MARL algorithms augmented with communication}. We provide several enhancements to our prior work \cite{Seraj2022LearningED}, including 1) the development of a novel Multi-Agent Heterogeneous Proximal Policy Optimization (MAH-PPO) algorithm to support sample-efficient learning of coordination policies for heterogeneous robots, 2) an architecture enhancement to support transferability to different environment sizes, resulting in non-parametricity in the number of agents, and 3) a noise-degradation channel with three different paradigms of loss. We find that HetNet can outperform baselines across several domains, scale to domain configurations in which the baselines fail to learn, and develop robustness to noise-degraded communication channels.

\vspace{-0.2cm}
\section*{Acknowledgments}
This work is supported by the Naval Research Lab under grant N00173-21-1-G009, Lockheed Martin Corporation under grant GR0-0000509, Office of Naval Research (ONR) under grant N00014-19-1-2076, and supported by the Laboratory Directed Research and Development program at Sandia National Laboratories.

Sandia National Laboratories is a multi-mission laboratory managed and operated by National Technology $\&$ Engineering Solutions of Sandia, LLC (NTESS), a wholly owned subsidiary of Honeywell International Inc., for the U.S. Department of Energy’s National Nuclear Security Administration (DOE/NNSA) under contract DE-NA0003525. This written work is authored by an employee of NTESS. The employee, not NTESS, owns the right, title and interest in and to the written work and is responsible for its contents. Any subjective views or opinions that might be expressed in the written work do not necessarily represent the views of the U.S. Government. The publisher acknowledges that the U.S. Government retains a non-exclusive, paid-up, irrevocable, world-wide license to publish or reproduce the published form of this written work or allow others to do so, for U.S. Government purposes. The DOE will provide public access to results of federally sponsored research in accordance with the DOE Public Access Plan.


%



\ifCLASSOPTIONcaptionsoff
  \newpage
\fi

\bibliographystyle{IEEEtran}
\bibliography{main}

%








\end{document}